%% file: main.tex
\documentclass[sigconf]{acmart}
\usepackage{graphicx} %
\usepackage{algorithm}
\usepackage{algorithmic} 
\usepackage{booktabs}
\usepackage{multirow}
\usepackage{graphicx}
\usepackage{subcaption}
\usepackage{tikz}
\usepackage{xcolor,pifont}
\usepackage{mdframed}
\usepackage{bbm}
\usepackage[shortlabels]{enumitem}
\usepackage[abbreviations]{foreign}

\usetikzlibrary{positioning, arrows.meta,backgrounds,shapes.geometric}

\definecolor{sageGreen}{rgb}{0.44, 0.5, 0.39}
\definecolor{beamerGreen}{rgb}{0.0, 0.6, 0.3}
\definecolor{lightbluegrey}{RGB}{176,196,222} 

\definecolor{sageGreen}{rgb}{0.44, 0.5, 0.39}
\definecolor{beamerGreen}{rgb}{0.0, 0.6, 0.3}
\definecolor{lightbluegrey}{RGB}{176,196,222} 

\newenvironment{takeaway}{%
  \par %
  \addvspace{\medskipamount} %
  \noindent %
  \begin{minipage}{\linewidth}%
  \begin{mdframed}[%
    linecolor=lightbluegrey,
    linewidth=1pt,
    topline=true,
    bottomline=true,
    leftline=true,
    rightline=true,
    backgroundcolor=lightbluegrey!20
  ]%
  \color{black}%
}{%
  \end{mdframed}%
  \end{minipage}%
  \par %
  \addvspace{\medskipamount} %
}

\usetikzlibrary{calc}

\title[Exposing the Vulnerability of Decentralized Learning to Membership Inference Attacks]{Exposing the Vulnerability of Decentralized Learning to Membership Inference Attacks Through the Lens of Graph Mixing
}
\date{\today}

\author{Ousmane Touat}
\email{ousmane.touat@insa-lyon.fr}
\affiliation{%
 \institution{CNRS, INSA Lyon - LIRIS}
 \city{Villeurbanne}
 \country{France}
}

\author{Jezekael Brunon}
\email{jezekael.brunon@insa-lyon.fr}
\affiliation{%
 \institution{INSA Lyon - LIRIS}
 \city{Villeurbanne}
 \country{France}
}

\author{Yacine Belal}
\email{yacine.belal@insa-lyon.fr}
\affiliation{%
 \institution{INSA Lyon - LIRIS}
 \city{Villeurbanne}
 \country{France}
}

\author{Julien Nicolas}
\email{julien.nicolas@insa-lyon.fr}
\affiliation{%
 \institution{CNRS, INSA Lyon - LIRIS, McGill - ILLS}
 \city{Villeurbanne}
 \country{France}
}

\author{C\'esar Sabater}
\email{cesar.sabater@insa-lyon.fr}
\affiliation{%
 \institution{CNRS, INSA Lyon - LIRIS}
 \city{Villeurbanne}
 \country{France}
}

\author{Mohamed Maouche}
\email{mohamed.maouche@inria.fr}
\affiliation{%
 \institution{Inria, INSA Lyon - CITI}
 \city{Villeurbanne}
 \country{France}
}

\author{Sonia Ben Mokhtar}
\email{sonia.ben-mokhtar@cnrs.fr}
\affiliation{%
 \institution{CNRS, INSA Lyon - LIRIS}
 \city{Villeurbanne}
 \country{France}
}

\input{content/abstract}

\input{content/macros.tex}



\begin{document}

\maketitle

\input{content/introduction}

\input{content/preliminaries}

\input{content/experimental_results}

\input{content/mixingrate}

\input{content/discussion-summary}

\input{content/related_works}

\input{content/conclusion}

\begin{acks}
This work was supported by the French government managed by the Agence Nationale de la Recherche (ANR) through France 2030 program with the reference ANR-23-PEIA-005 (REDEEM project).
\end{acks}

\bibliographystyle{ACM-Reference-Format}
\bibliography{bibliography}

\end{document}

%% file: content/abstract.tex
\begin{abstract}
The primary promise of decentralized learning is to allow users to engage in the training of machine learning models in a collaborative manner while keeping their data on their premises and without relying on any central entity. However, this paradigm necessitates the exchange of model parameters or gradients between peers. Such exchanges can be exploited to infer sensitive information about training data, which is achieved through privacy attacks (\eg Membership Inference Attacks -- MIA).
In order to devise effective defense mechanisms, it is important to understand the factors that increase/reduce the vulnerability of a given decentralized learning architecture to MIA. 
In this study, we extensively explore the vulnerability to MIA of various decentralized learning architectures by varying the graph structure (\eg number of neighbors), the graph dynamics, and the aggregation strategy, across diverse datasets and data distributions.
Our key finding, which to the best of our knowledge we are the first to report, is that the vulnerability to MIA is heavily correlated to (i) the local model mixing strategy performed by each node upon reception of models from neighboring nodes and (ii) the global mixing properties of the communication graph. We illustrate these results experimentally using four datasets and by theoretically analyzing the mixing properties of various decentralized architectures. \revision{We also empirically show that enhancing mixing properties is highly beneficial when combined with other privacy-preserving techniques such as Differential Privacy.} Our paper draws a set of lessons learned for devising decentralized learning systems that reduce by design the vulnerability to MIA. 
  
\end{abstract}

%% file: content/macros.tex
\newcommand{\modelVector}{\theta}
\newcommand{\modelVectorIt}[1]{\modelVector^{(#1)}}
\newcommand{\model}[2]{\modelVectorIt{#1}_{#2}}
\newcommand{\iter}{t}
\newcommand{\mixMatrix}{W}
\newcommand{\mixMatrixIt}[1]{\mixMatrix^{(#1)}}
\newcommand{\iterCnt}{T}
\newcommand{\mixMatrixItEl}[3]{\mixMatrixIt{#1}_{#2,#3}}
\newcommand{\oneVec}{\mathbbm{1}}
\newcommand{\modelSymb}{\theta} 
\newcommand{\globalModel}{\modelSymb^\star}
\newcommand{\graph}[1]{G^{(#1)}}
\newcommand{\edges}[1]{E^{(#1)}}

\newcommand{\lTwoNorm}[1]{\|#1\|_2}
\newcommand{\secondEig}[1]{\lambda_2\left(#1\right)}
\newcommand{\mixSequence}{\mixMatrix^\star}

\newcommand{\dataset}[1]{\mathcal{D}_{#1}}
\newcommand{\globalDataset}{\mathcal{D}}
\newcommand{\loss}[3]{l_{#1}(#2,#3)}
\newcommand{\gradient}[3]{\nabla \loss{#1}{#2}{#3}}
\newcommand{\learningRate}{\eta}
\newcommand{\neighbors}[2]{\mathcal{N}^{(#1)}_{#2}}
\newcommand{\neighSize}{k}
\newcommand{\vect}{\modelVector}
\newcommand{\avgVec}{\tilde{\vect}}
\newcommand{\partySet}{\mathcal{V}}
\newcommand{\samo}{SAMO}
\newcommand{\tpratlowfpr}{TPR@1\%FPR}

\newcommand{\peerTime}{p}

\newcommand{\incomingModelsVar}[1]{\Theta_{#1}}
\newcommand{\modelASynch}[1]{\modelSymb_{#1}}
\newcommand{\sample}{x}
\newcommand{\sampleDom}{\mathcal{X}} 
\newcommand{\modelDom}{\mathcal{M}}
\newcommand{\partsOf}[1]{\wp(#1)}
\newcommand{\attribDom}{\mathcal{Z}}
\newcommand{\labelDom}{\mathcal{Y}}
\newcommand{\predictor}{P}
\newcommand{\mpEntropy}[2]{M_{\text{MPE}}(#1,#2)}
\newcommand{\attack}[2]{\attackSymb(#1,#2)}
\newcommand{\modelPredDist}[2]{#1[#2]}
\newcommand{\attackSymb}{\mathcal{A}}
\newcommand{\attackThreshold}{\tilde{\tau}}
\newcommand{\attackMPESymb}{\attackSymb_{\text{MPE}}}
\newcommand{\attackMPE}[3]{\attackMPESymb(#1,#2,#3)}
\newcommand{\testDataSet}[1]{\dataset{#1}^{t}}

\newcommand{\revision}[1]{\textcolor{black}{#1}}

%% file: content/introduction.tex
\section{Introduction}

Federated learning~\cite{fedavg}, which allows training machine learning models while keeping users' data on their premises has gained significant interest in recent years both from academia and industry~\citep{hard2018federated,banabilah2022federated, wang2019adaptive, nicolas2024differentially}. However, this approach is often criticized because of the trust it requires on the aggregation server, which may be subject to failures~\citep{kairouz2021advances}, attacks~\citep{bhagoji2019analyzing}, scalability issues~\citep{li2020federated}, \etc.
In order to deal with these challenges, Decentralized Learning, which builds on the peer-to-peer exchange of model parameters between participants is gaining an increasing focus (\eg,~\cite{lian2017,ormandi2013, Hegeds2019GossipLA}). Indeed, decentralized architectures have been heavily studied for decades and their scalability and resilience properties have been extensively documented both theoretically and experimentally in the literature~\citep{baran1964distributed, lynch1996distributed, van2003astrolabe}. This makes decentralized learning a natural avenue to explore for large-scale, distributed learning tasks~\citep{lian2017can, hendrikx2019accelerated, wu2024deep}.

However, whether federated or decentralized, collaborative learning has been shown to be vulnerable to privacy attacks and in particular to Membership Inference Attacks (MIA)~\cite{nasr2019comprehensive, melis2019exploiting,Zhang2020}. This family of attacks aims at predicting whether a given data item has been used or not in the model training set, which can reveal sensitive information about the data owner (\eg a model trained to predict the risk of heart disease, could reveal sensitive information such as a patient's history of cholesterol levels, blood pressure, or family medical history~\citep{korenmembership}). Moreover, since MIA is closely related to other privacy attacks~\cite{yeomPrivacyRiskMachine2018a}, the assessment of the vulnerability against such attack provides valuable insight into a model's overall vulnerability to other privacy attacks~\cite{pasquini_security_2023}. 

In this context, an active debate started in the research community to understand whether decentralization increases or decreases the vulnerability to MIA compared to FL architectures. For instance, \citet{pasquini_security_2023} points out the vulnerability of decentralized topologies, while \citet{ji2024re} contradicts this study by claiming that FL topologies are weaker. While these papers contribute to advancing the global knowledge on this topic, these studies are limited because they mainly focus on static graph topologies, ignoring decades of research on dynamic peer-to-peer systems enabled by their underlying random peer sampling protocols (RPS)~\citep{jelasity2007gossip, voulgaris2005cyclon}. 

In this paper, we aim to contribute to this debate by analyzing the key factors that make decentralized architectures more or less vulnerable to MIA. Indeed, there are various ways of deploying a decentralized learning task, which differ according to the graph topology \revision{(\eg, a $k$-regular graph where each node's view size, or number of neighbors, is consistently $k$)}, graph dynamics (\eg static graph, vs graph built over RPS protocols), communication synchronicity (synchronous vs asynchronous model exchange) and model mixing and dissemination strategy. 

In this study, we focus on the MIA vulnerability of \revision{peer-to-peer based} learning over asynchronous, $k$-regular graphs (also defined as Gossip Learning in~\cite{hegedHus2019gossip,ormandi2013}) and vary all the remaining factors. Specifically, (i) we explore two local model mixing and dissemination strategies by introducing the \textit{send-all-merge-once (SAMO)} strategy; (ii) we vary graph dynamics; (iii) we vary graph density through the number of neighbors. Beyond these key architectural features, we further consider four commonly used datasets and corresponding models and study the impact of i.i.d. and non-i.i.d. data distributions. 
We evaluate the MIA vulnerability of the overall decentralized architecture by considering that the attacker can observe all the messages exchanged in the system and run the MIA against each participant using messages received from its neighbors.

Our key finding is that MIA vulnerability is heavily correlated with two factors: (i) the \emph{local model mixing} and dissemination strategy, which determines how a node aggregates models received from his neighbors and how he disseminates his updated model and (ii) the \emph{global mixing properties of the communication graph}. For instance, we show that a static highly connected graph has a similar vulnerability to a dynamic graph of the same size. Instead, in weakly connected graphs (\eg 2-regular graphs), graph dynamics brought by RPS protocols clearly decrease MIA vulnerability. Beyond local and global mixing properties, we also observe that learning tasks over non-i.i.d. data distributions clearly increase MIA vulnerability.

Summarizing, our paper makes the following contributions:
\begin{itemize}

\item \textbf{Comprehensive Study of MIA Vulnerabilities in Decentralized Learning}:
To the best of our knowledge, we propose the first study that scrutinizes how MIA vulnerability is affected by decentralized learning settings, focusing on the roles of graph connectivity, graph dynamics, model mixing, and data distributions. 

\item \textbf{Insights into Mixing Properties and Privacy}:
We identify that both local and global model mixing properties are critical determinants of MIA vulnerability. Our findings reveal how dynamic graph settings and enhanced view sizes improve model mixing and mitigate privacy risks. Note that all our results can be reproduced using the code available in the following public repository~\footnote{https://gitlab.liris.cnrs.fr/otouat/exposingvulnerabilitydl}.

\item \textbf{Recommendations for Privacy-Aware Decentralized Learning beyond generalization error}:
Based on our findings, we provide design recommendations for decentralized learning systems that aim to balance utility with reduced MIA vulnerabilities, emphasizing the importance of dynamics and robust mixing protocols. We show that generalization error, which has been highlighted by previous works~\cite{pasquini_security_2023}, is not the only key factor in determining the privacy risk. 

\end{itemize}

The remainder of this paper is structured as follows: we first present preliminaries on decentralized learning, gossip learning, and a variant of gossip learning we consider in this paper as well as background on MIA and our threat model in Section~\ref{sec:preliminaries}. We then present our experimental analysis in Section~\ref{sec:exp}. We complement our study with a theoretical analysis of graph mixing properties in Section~\ref{sec:mixing}. Then, we summarize our actionable recommendations in Section~\ref{sec:summary}. Finally, we discuss related research efforts in Section~\ref{sec:related} before concluding the paper in Section~\ref{sec:conclusion}.

%% file: content/preliminaries.tex
\section{Preliminaries and Setup}\label{sec:preliminaries}

In this section we present an overview of decentralized learning (Section~\ref{subsec:dl}) and the two learning protocols we consider, Base Gossip Learning (Section~\ref{subsec:gl}) and a variant of this protocol (Section~\ref{subsec:samo}). We detail the neighbor selection strategy for these two protocols in Section~\ref{subsec:rps}. Then, we introduce membership inference attacks (Section~\ref{subsec:mia}) and our considered threat model (Section~\ref{subsec:threatmodel}). 

\subsection{Decentralized Learning}
\label{subsec:dl}

Decentralized Learning refers to a set $\partySet = \{1, \dots, n\}$ of nodes, which aim to collaboratively train a machine learning model without relying on a central parameter server. Given a set of attributes $\attribDom$, a set of labels $\labelDom$  and  a model domain $\modelDom$, each node  $i \in \partySet$  has its local dataset denoted as $\dataset{i} \subset \sampleDom = \attribDom \times \labelDom$ and their joint goal is to learn a global model $\globalModel \in \modelDom$ which minimizes the loss function:
\begin{equation}
	\mathcal{L}(\globalModel) = \frac{1}{n}  \sum \limits_{\underset{}{i = 1}}^{n} \loss{i}{\globalModel}{\dataset{i}},
\end{equation}
where $\loss{i}{\cdot}{\cdot}: \modelDom \times \partsOf{\sampleDom} \to \mathbb{R}$  is the local loss function \footnote{for a set $S$ we denote by $\partsOf{S}$ to the set of all subsets of $S$} of node $i$ and  $\globalDataset = \bigcup_{i \in \partySet} \dataset{i}$ as defined in~\citep{bellet2017fast}. 

In decentralized learning, each node performs three types of operations over models: local model updates, model exchanges, and model aggregation as described below.

\paragraph{\textbf{Local Updates}} 
Each node $i \in \partySet$ updates its current model $\modelASynch{i}$  using its local training dataset $\dataset{i}$ as follows:
\begin{equation}
\modelASynch{i} = \modelASynch{i}  - \learningRate \gradient{i}{\modelASynch{i}}{\dataset{i}},
\end{equation}
where $\learningRate$ is the learning rate and  $\gradient{i}{\modelASynch{i}}{\dataset{i}}$ is the gradient of $\loss{i}{\modelASynch{i}}{\dataset{i}}$.

As each node may update its model parameters by performing more than one local step, we will slightly abuse notation and still note  $\learningRate\gradient{i}{\modelASynch{i}}{\dataset{i}}$ the sum of gradients obtained by sequentially applying a previously defined number of local steps with a given learning rate $\learningRate$.   

\paragraph{\textbf{Model Exchange}} 
Each node $i \in \partySet$ exchanges its current model with one or many other nodes in its neighborhood. This neighborhood can be static or dynamic as further discussed in Section~\ref{subsec:rps}. Let $\peerTime \in \mathbb{N}$ be our time parameter, $\neighbors{\peerTime}{i} \subset \partySet$ is the view of party $i$ at time $\peerTime$. In practice, parties obtain their view through a random peer sampling service (RPS)~\citep{jelasity2007gossip, devosEpidemicLearningBoosting2023}. 
At each time $\peerTime \in \mathbb{N}$, the sets of views of all nodes $\{ \neighbors{\peerTime}{i} \}_{i \in \partySet}$ define a graph $\graph{\peerTime}=(\partySet, \edges{\peerTime} \subset \partySet \times \partySet)$ where an edge $(i,j) \in \edges{\peerTime} \iff j \in \neighbors{\peerTime}{i}$.

\paragraph{\textbf{Local Aggregation}} To gain information from their neighbors, nodes aggregate their current model with received models. While for different purposes there exist many alternatives to perform this aggregation~\cite{qi2024model, li2021model, karimireddy2020scaffold}, we consider simple averaging of the current incoming models as in \cite{liDecentralizedFederatedLearning2022}.

In the following section, we precisely describe how these operations are performed in the protocols we study.

\subsection{Base Gossip Learning}\label{subsec:gl} 

In this paper, we focus on a particular type of decentralized learning, known as Gossip learning (GL)~\cite{hegedHus2019gossip,ormandi2013}, which assumes that nodes behave asynchronously. In Gossip Learning, each node initially starts with a common model $\modelASynch{0}$ and a set of neighbors also referred to as a view, and periodically and independently wakes up to perform the above three operations \ie local updates, model exchange, and local aggregation.

Specifically, the GL algorithm defined in~\cite{ormandi2013} is presented in Algorithm~\ref{alg:base}. In this algorithm, each node starts with a shared initial model $\modelASynch{0}$ (line 1). When a node $i \in \partySet$ wakes up at time $\peerTime \in \mathbb{N}$, it selects a random neighbor from $\neighbors{\peerTime}{i}$ and sends its current model $\modelASynch{i}$ (lines \ref{alg:base.exchange.1} and \ref{alg:base.exchange.2}). When it receives a model $\modelASynch{j}$, it aggregates it with its current model  $\modelASynch{i}$ (line \ref{alg:base.aggregate}) and performs a local update (line \ref{alg:base.localupdate}). 

\begin{figure}[ht]
\includegraphics[width=8cm]{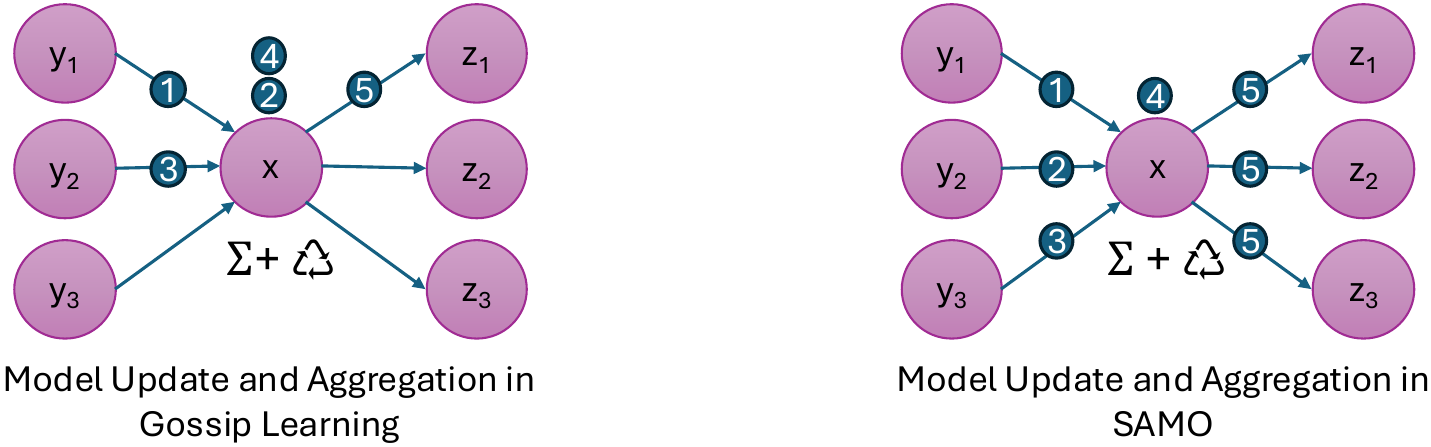}
\caption{Model Update and Aggregation in Gossip Learning (left) vs in SAMO (right)}\label{fig:gossipvssamo}
\end{figure}

This algorithm is illustrated with an example depicted in the left part of Figure~\ref{fig:gossipvssamo}. In this figure, node $x$ receives a model from node $y_1$ (step \ding{182}), then performs a local update after aggregating the two models (step \ding{183}). Later, it receives a model from node $y_2$ (step \ding{184}) and performs again model aggregation and local update (step \ding{185}). Finally, upon waking up, it randomly selects one of its neighbors, \ie node $z_1$ to which it sends its latest model (step \ding{186}).

We observe in this algorithm that models are aggregated in a pairwise manner, after which a local update is performed by the model owner. Furthermore, we observe that the model dissemination strategy is slow as each node sends its updated model to only one randomly selected neighbor. To increase model mixing, we investigate in this paper an alternative to the classical gossip learning algorithm, which we present in the next section. 

\begin{algorithm}[ht]
	\caption{Base GL Protocol \label{alg:base}}
	\begin{algorithmic}[1]
		 \STATE{\textbf{Input:} Communication topologies $\neighbors{\peerTime}{i}$ for each time $p$ and node $i \in \partySet$;  initial model $\modelSymb_{i} = \modelSymb_0$, local data-sets $\dataset{i}$, loss function $\loss{i}{\cdot}{\cdot}$ for all $i \in \partySet$; learning rate $\learningRate$} 
	 \FOR{each node $i \in \partySet$ that \textbf{wakes up} within  time frame $\peerTime$}
	  \STATE{Select a node $j \in \neighbors{\peerTime}{i}$ at random \label{alg:base.exchange.1}}
	  \STATE{Send $\modelSymb_i$ to node $j$ \label{alg:base.exchange.2}}
	\ENDFOR

	\FOR{each node  $i \in \partySet$ \textbf{that receives a model} $\modelSymb_j$:}
	 \STATE{Set $\modelSymb_i \gets \frac{\modelSymb_i + \modelSymb_j}{2}$ \label{alg:base.aggregate}}  \qquad \qquad  \COMMENT{aggregate with incoming model}
	 \STATE{Set $\modelSymb_i \gets \modelSymb_i - \learningRate\gradient{i}{\modelSymb_i}{\dataset{i}}$}\quad \COMMENT{perform local updates \label{alg:base.localupdate}}
	\ENDFOR 
	\end{algorithmic}
\end{algorithm}

\subsection{Send-All-Merge-Once Protocol}
\label{subsec:samo} 
We hereby introduce in Algorithm \ref{alg:samo} a variant of the GL protocol, called \emph{Send-All-Merge-Once} (\samo{}) that is designed to improve the speed of model mixing over time.
In \samo, when a node receives a model, it stores it (as shown in line \ref{alg:samo.store}) instead of directly aggregating it and performing local updates. 
Later, when a node wakes up, if models were received, it aggregates them with its own (line \ref{alg:samo.aggregate}) and performs local updates (line \ref{alg:samo.localupdates}). Finally, it sends its current model to all its neighbors (line \ref{alg:samo.exchange}).
This algorithm is illustrated with an example depicted in the right part of Figure~\ref{fig:gossipvssamo}. In this figure, node $x$ receives models from its neighbors $y_1$, $y_2$ and $y_3$ and stores them locally (steps \ding{182}, \ding{183} and \ding{184}). Later, upon waking up, node $x$ aggregates these models with its local model performs local updates (step \ding{185}), and sends the new model to all its neighbors (step \ding{186}). 

As exchanges involve all the neighbors of a node, \samo{} has a higher communication cost than the Base GL Protocol. Nevertheless, it better resists MIA (see Section~\ref{sec:exp}).%

\begin{algorithm}[ht]
	\caption{SAMO Protocol}
	\begin{algorithmic}[1]
		  \STATE{ \textbf{Input:} Communication topologies $\neighbors{\peerTime}{i}$ for each time frame $p$ and node $i \in \partySet$,  initial model $\modelSymb_{i} = \modelSymb_0$, local datasets $\dataset{i}$, incoming model sets $\incomingModelsVar{i} = \{\modelSymb_0\}$ and loss function $\loss{i}{\cdot}{\cdot}$ for all $i \in \partySet$;   learning rate $\learningRate$}
		\FOR{each node $i \in \partySet$ that \textbf{wakes up} within time frame $\peerTime$} 
            \IF{$|\incomingModelsVar{i}|> 1$} 
            \STATE{Set $\modelSymb_i \gets \frac{1}{|\incomingModelsVar{i}|} \sum_{\modelSymb \in \incomingModelsVar{i} } \modelSymb $ \label{alg:samo.aggregate}} \COMMENT{aggregate models} 
			\STATE{Set $\modelSymb_i  \gets \modelSymb_i - \learningRate \gradient{i}{\modelSymb_i}{\dataset{i}}$\label{alg:samo.localupdates}} \COMMENT{ do local updates }
				\STATE{Set $\incomingModelsVar{i} \gets \{ \modelSymb_i \}$}
            \ENDIF
			\STATE{Send $\modelSymb_i$ to each node $j \in \neighbors{\peerTime}{i}$ \label{alg:samo.exchange}}
		\ENDFOR 
		\FOR{each node  $i \in \partySet$ \textbf{that receives a model} $\modelSymb_j$}
		\STATE{Set $\incomingModelsVar{i} \gets \incomingModelsVar{i} \cup  \modelSymb_j$ \label{alg:samo.store}} 
		\ENDFOR 
	\end{algorithmic}
	\label{alg:samo} 
\end{algorithm}

\subsection{Peer Sampling Service} \label{subsec:rps}

In this paper, we aim to study both static and dynamic topologies. For both GL and \samo{}, we assume that nodes rely on a Random Peer Sampling Service such as~\cite{jelasity2007gossip}.  Specifically, in both protocols, neighbors of each node $i \in \partySet$, are chosen as follows. Initially, $\neighbors{1}{i}$ is determined by a $k$-regular graph $\graph{1}$ chosen at random by the peer sampling service. In the static setting, neighbors do not change anymore. In the dynamic setting we adopt the PeerSwap method~\cite{guerraoui2024peerswap}: when node $i$ wakes up at time $\peerTime$, it swaps the view with a randomly selected neighbor before doing anything else. More precisely, $\neighbors{\peerTime}{i}$ is set to $(\neighbors{\peerTime-1}{j}\setminus \{i\}) \cup \{j\}$ through the peer sampling service, which also updates
\begin{align*} 
\neighbors{\peerTime}{j} &\gets  (\neighbors{\peerTime-1}{i}  \setminus \{j\}) \cup \{i\} \\
 \neighbors{\peerTime}{k} &\gets (\neighbors{\peerTime-1}{k} \setminus \{i\}) \cup \{j\} \text{ for all }k \in \neighbors{\peerTime-1}{i} \setminus \{j\} \\ 
 \neighbors{\peerTime}{k} &\gets (\neighbors{\peerTime-1}{k} \setminus \{j\}) \cup \{i\} \text{ for all }k \in \neighbors{\peerTime-1}{j}\setminus \{i\}
 \end{align*} 
to keep views consistent. The swap is equivalent to changing the position of $i$ and $j$ in the current graph, which remains $k$-regular.

\subsection{Membership Inference Attack}
\label{subsec:mia} 

A Membership Inference Attack (MIA) is a privacy attack whose goal is to infer for a given model $\modelSymb \in \modelDom$ whether or not a sample $\sample \in \sampleDom$ belongs to the data used to train this model~\cite{shokri2017membership}. More formally, it is an algorithm $\attackSymb : \modelDom \times \sampleDom \to \{0,1\}$ such that $\attack{\modelSymb}{\sample}=1$ if it predicts that $\sample$ belongs to the training dataset of model $\modelSymb$ and $\attack{\modelSymb}{\sample}=0$ otherwise. 

Many MIAs have been proposed and range from expensive approaches that train ML models to predict membership such as neural shadow models \cite{shokri2017membership} to information-theoretic estimators that leverage prediction entropy or confidence~\cite{salem2019,songSystematicEvaluationPrivacy2020}. 
In our work, we use the Modified Prediction Entropy Attack~\cite{songSystematicEvaluationPrivacy2020}. We first explain the Modified Prediction Entropy (MPE) measure and then the attack.

Let $\predictor: \labelDom \to [0,1]$ be a probability distribution over $\labelDom$ and $y \in  \labelDom$. The MPE measure $\mpEntropy{\cdot}{\cdot}$ is defined as follows:  
\begin{align}
	\mpEntropy{\predictor}{y} = & 
	- (1 - \predictor(y)) \log (\predictor(y)) \notag \\
	& - \sum_{y' \in \labelDom \setminus \{y\}} \predictor(y') 
	\log (1 - \predictor(y')).
\end{align}
The MPE Attack used to discriminate if the sample $(z, y) \in \sampleDom$ was used to train model $\modelSymb \in \modelDom$ is defined as follows:
\begin{equation}
	\attackMPE{\modelSymb}{(z, y)}{\revision{\attackThreshold}} = 
	\begin{cases} 
		1 & \text{if } \mpEntropy{\modelPredDist{\modelSymb}{z}}{y} \leq \attackThreshold \\
		0 & \text{otherwise}
	\end{cases}.
\end{equation}
where $\modelPredDist{\modelSymb}{z}: \labelDom \to [0,1]$ is the probability distribution that model $\modelSymb$ assigns to each label in $\labelDom$ given input attributes $z \in \attribDom$ and $\attackThreshold$ is the attack threshold computed using training dataset $\dataset{i}$ of a target node (or victim) $i \in 
\partySet$ as well as a test dataset $\testDataSet{i}$ that was not used to train $\modelSymb$. 
The idea is that the prediction entropy can discriminate between member and non-member data, as the target model $\modelSymb$ has a lower prediction entropy on train data.

It is unrealistic to deploy $\attackMPESymb$ in practice as it uses the training dataset of the victim to get the optimal threshold $\attackThreshold$. However, this approach offers an upper bound on the worst-case attacker whose information is the probability distributions $\{ \modelPredDist{\modelSymb}{z} \}_{z \in \attribDom}$ generated by the target model $\modelSymb$ and sample $\sample$. Therefore it is an informative privacy assessment metric.  Moreover, it is significantly more practical to deploy than machine learning-based discriminators.

\subsection{Threat Model}
\label{subsec:threatmodel} 
As an attacker, we consider an omniscient observer that at regular time intervals recovers the current models of all nodes $\{ \modelASynch{i} \}_{i \in \partySet}$ and performs $\attackMPESymb$ on each one of them, targeting each data sample of each node. 
It allows us to evaluate the vulnerability of all nodes in the system wherever their position is in the communication graph.
This attacker is similar to the passive attacker defined in~\cite{pasquini_security_2023} to assess the vulnerability of static decentralized learning against MIA.
\revision{We do not consider other attack models that consider actively malicious nodes that can deviate from the protocol to gain information~\cite{pasquini_security_2023}.}
Finally, a more detailed explanation of the metrics we used to compute the vulnerability to MIA is provided in Section \ref{sec:exp.metrics}.

%% file: content/experimental_results.tex
\section{Experimental Study}

\label{sec:exp} 

In this section, we present the findings of our study, which investigates the factors influencing the tradeoff between utility and privacy in Gossip learning. Specifically, we address the following research questions:

\begin{itemize}
\item \textbf{RQ1}: What is the impact of using the SAMO protocol compared to Base Gossip?
\item \textbf{RQ2}: Is there an advantage to using a dynamic setting compared to a static one?
\item \textbf{RQ3}: Viewed through the lens of canary-based evaluation (probing worst-case memorization), what advantages does using a dynamic setting offer?
\item \textbf{RQ4}: What is the impact of the view size on both static and dynamic settings?
\item \textbf{RQ5}: What is the impact of non-i.i.d. data distribution?
\item \textbf{RQ6}: Is the link between generalization error and MIA vulnerability maintained in our studied setting?
\item \textbf{RQ7}: How using a dynamic setting can complement other privacy-preserving techniques such as Differential Privacy?
\end{itemize}

\subsection{Experimental Setup \& Configurations}
\paragraph{\textbf{Implementation of Gossip Learning Protocol}} We consider two gossip learning protocols:  the Base Gossip protocol and  \samo{}. Both protocols are implemented using our fork of the GossiPy python framework~\cite{Mirko_2024} with Python 3.11 and Pytorch. 

To simulate the asynchronous nature of the gossip protocol, the execution is divided into discrete time units called ticks. Each round of communication is represented by 100 ticks and each node $i \in \partySet$ waits $\Delta_{i}$ ticks between wake-ups. The waiting time  $\Delta_{i}$ is sampled from a normal distribution $\mathcal{N}(\mu,\,\sigma^2)$ with $\mu = 100$ and $\sigma^2 = 100$ at the beginning of the execution. 

\label{par:roundDefinition}
\paragraph{\textbf{Communication Topology}} In our setup we simulate 150 nodes that are connected together in an initial  $k$-regular graph topology with the view size $k \in \{2, 5, 10, 25\}$. We consider both static and dynamic settings.

\paragraph{\textbf{Datasets and Partitioning}}
The efficacy of the proposed protocol is assessed on a number of datasets described in Table~\ref{tab:dataset-characteristics}.
For experiments on independent and identically distributed (i.i.d.) data, we utilize standard image classification datasets: CIFAR-10, CIFAR-100~\cite{krizhevskyLearningMultipleLayers2009} and FashionMNIST~\cite{fashionMNIST}, alongside the Purchase100 dataset~\cite{shokri2017membership}, which is commonly used in privacy attack research for tabular data.
These datasets represent established benchmarks within the decentralized learning and privacy-focused research communities.
In this configuration, the training data is distributed uniformly across all nodes in an equal partition. 
Specifically, this partitioning involves sampling each node's training data i.i.d. from the base dataset's training split. Furthermore, each node's respective test data is also sampled i.i.d. directly from this same base training split. The global test set (originating from the base dataset's distinct test split) is maintained separately to assess the global test accuracy of each model.

\begin{table*}[t]
\caption{Dataset Characteristics}
\label{tab:dataset-characteristics}
\begin{tabular}{l c c c c c p{0.25\textwidth}}
\toprule
\textbf{Dataset} & \textbf{Train Set} & \textbf{Test Set} & \textbf{Input Size} & \textbf{Classes} & \textbf{Model} & \textbf{Description} \\
\midrule
CIFAR-10 & 50,000 & 10,000 & $32 \times 32 \times 3$ & 10 & CNN & Color images across 10 classes including animals, vehicles \\
CIFAR-100 & 50,000 & 10,000 & $32 \times 32 \times 3$ & 100 & ResNet-8 & Fine-grained color images with 100 object classes \\
Fashion-MNIST & 60,000 & 10,000 & $28 \times 28 \times 1$ & 10 & CNN & Grayscale images of clothing and fashion accessories \\
Purchase-100 & 157,859 & 39,465 & 600 & 100 & MLP & A tabular dataset of customer purchases to classify buying behavior \\
\bottomrule
\end{tabular}
\end{table*}

\paragraph{\textbf{Training Setup}}

For our experiments, we trained each model with Gossip Learning (base algorithm and our variant), varying hyperparameters such as the learning rate, the number of local steps, or the used model for the dataset depending on its task complexity. Table~\ref{tab:training-setup} shows the training setup for all our datasets, including the used model and the number of parameters. For light image datasets such as CIFAR-10 or Fashion-MNIST, we use a light CNN architecture. For Purchase100 we follow the implementation form Nasr et al.~\cite{nasr2019comprehensive}, by using a 4-layer fully-connected neural network. For more challenging image datasets such as CIFAR-100, we opt to use ResNet models~\cite{he2016residual}. We initialize each node's model using Kaiming normal initialization function~\cite{he2015}.

\begin{table*}[t]
\caption{Training Configuration}
\label{tab:training-setup}
\begin{tabular}{l c c c c c c c}
\toprule
\textbf{Dataset} & \textbf{Model} & \textbf{Parameters} & \textbf{Learning Rate} & \textbf{Momentum} & \textbf{Weight Decay} & \textbf{Local Epochs} & \textbf{Rounds}\\
\midrule
CIFAR-10 & CNN & 124k & 0.01 & 0 & 5e-4 & 3 & 250 \\
CIFAR-100 & ResNet-8 & 1.2M & 0.001 & 0.9 & 5e-4 & 5 & 500 \\
Fashion-MNIST & CNN & 124k & 0.01 & 0.9 & 5e-4 & 3 & 250 \\
Purchase-100 & MLP & 1.3M & 0.01 & 0.9 & 5e-4 & 10 & 250 \\
\bottomrule
\end{tabular}
\end{table*}

\subsection{Metrics}
\label{sec:exp.metrics} 
\paragraph{\textbf{Utility: Global Test Accuracy}}
In all experiments, we assess the utility of the node models using top-1 accuracy on the global test set.
For a model $\theta$, we define this metric as follows:
\begin{equation}
    \text{acc}(\theta, \mathcal{D_{\text{test}}}) = \frac{1}{|\mathcal{D_{\text{test}}}|} \sum_{(z,y) \in \mathcal{D_{\text{test}}}} \mathbb{I}\bigl(\arg\max \theta(z) = y\bigr),
\end{equation}
where $\mathbb{I}$ is the indicator function.
\paragraph{\textbf{Privacy: MIA Accuracy}}
We measure the accuracy of the MIA attack. Let $\theta$ be the victim model, $\attackSymb_{\text{MPE}}$ be the attack described in Section~\ref{subsec:mia} and $\mathcal{D}_{att} \subset \attribDom \times \labelDom \times \{0,1\}$ a set of member and non-member data. Membership in $\mathcal{D}_{att}$ is defined with label $m$ (1 for member and 0 for non-member) and usually sampled equally from the local train and test set respectively.  We then define the MIA vulnerability as below:
\begin{equation}
    \text{acc}_{\attackSymb}(\theta, \mathcal{D}_{att}) = \frac{1}{|\mathcal{D}_{att}|} \sum_{(z,y,m) \in \mathcal{D}_{att}} \mathbb{I}\bigl(\attackMPE{\modelSymb}{(z, y)}{\attackThreshold} = m),
\end{equation}
The threshold $\attackThreshold$ is implicitly defined as the value-maximizing the attack accuracy.

\paragraph{\textbf{Privacy: MIA \tpratlowfpr}} We also compute this metric, using the MPE scores, $\mpEntropy{\modelPredDist{\modelSymb}{z}}{y}$, calculated for all samples in $\mathcal{D}_{att}$. 
By varying the decision threshold applied to these MPE scores (a lower score indicates member), we can generate a Receiver Operating Characteristic (ROC) curve, plotting the True Positive Rate (TPR) against the False Positive Rate (FPR). 
The \tpratlowfpr is the value of the TPR obtained from the ROC curve at the point where the FPR is less than or equal to 1\%.
\begin{multline}
    \label{eq:tpr_at_fpr_via_roc}
    \text{\tpratlowfpr}(\theta, \mathcal{D}_{att}) = \\
    \frac{1}{|\mathcal{D}_{att, m=1}|} \sum_{(z,y) \in \mathcal{D}_{att}, m=1} \mathbb{I}\bigl(\attackMPE{\modelSymb}{(z, y)}{\attackThreshold_{1\%}} = 1)
\end{multline}

The threshold $\attackThreshold_{1\%}$ is implicitly defined as the value yielding the point on the ROC curve where FPR $\le 1\%$.
\paragraph{\textbf{Generalization error}}
We define a generalization error metric as the gap between the local train accuracy and the local test accuracy. 
Formally, for node $i$ that trains model $\theta_i$ with its local dataset $\mathcal{D}_{i}$, we define the generalization error as below: 
\begin{equation}
\label{eq:gen_error}
    \text{gen\_error}(\theta_i,\mathcal{D}_{i}) = \text{acc}(\theta_i, \mathcal{D}_{i,\text{train}}) - \text{acc}(\theta_i, \mathcal{D}_{i,\text{test}}),
\end{equation}
with $\mathcal{D}_{i,\text{train}}$ and $\mathcal{D}_{i,\text{test}}$ being respectively the local training and test dataset splits of node $i$.

At the end of each round of communication (defined in section~\ref{par:roundDefinition} ), we record measurements on the model of each node and subsequently report the mean value aggregated across the nodes.

\subsection{Comparing SAMO to Base Gossip (RQ1)}
\begin{figure}[ht]
    \begin{minipage}{\columnwidth}
    \centering
    \includegraphics[width=0.5\textwidth]{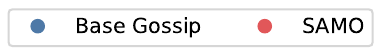}
    \\
    \includegraphics[width=0.90\columnwidth]{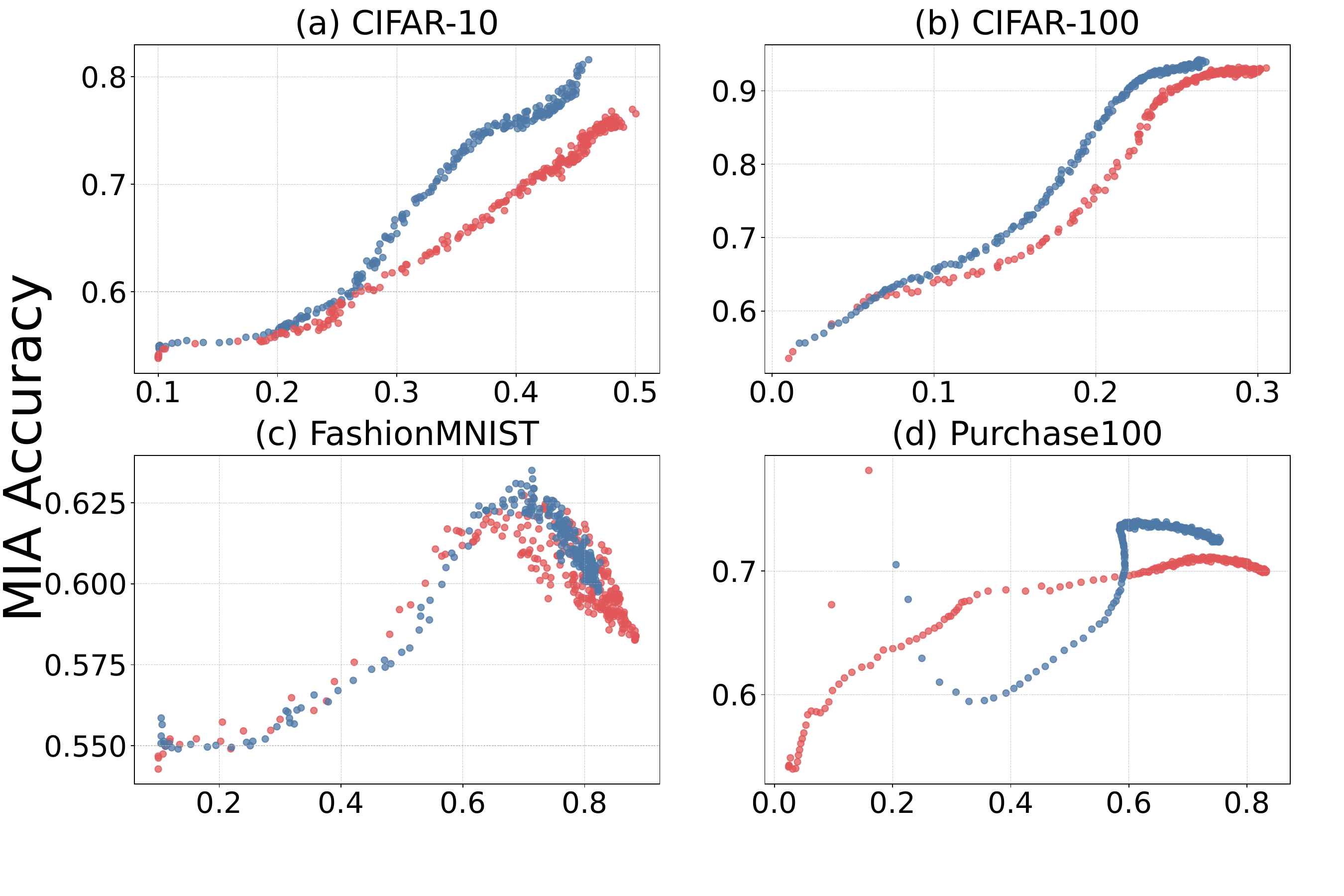}
    \includegraphics[width=0.90\columnwidth]{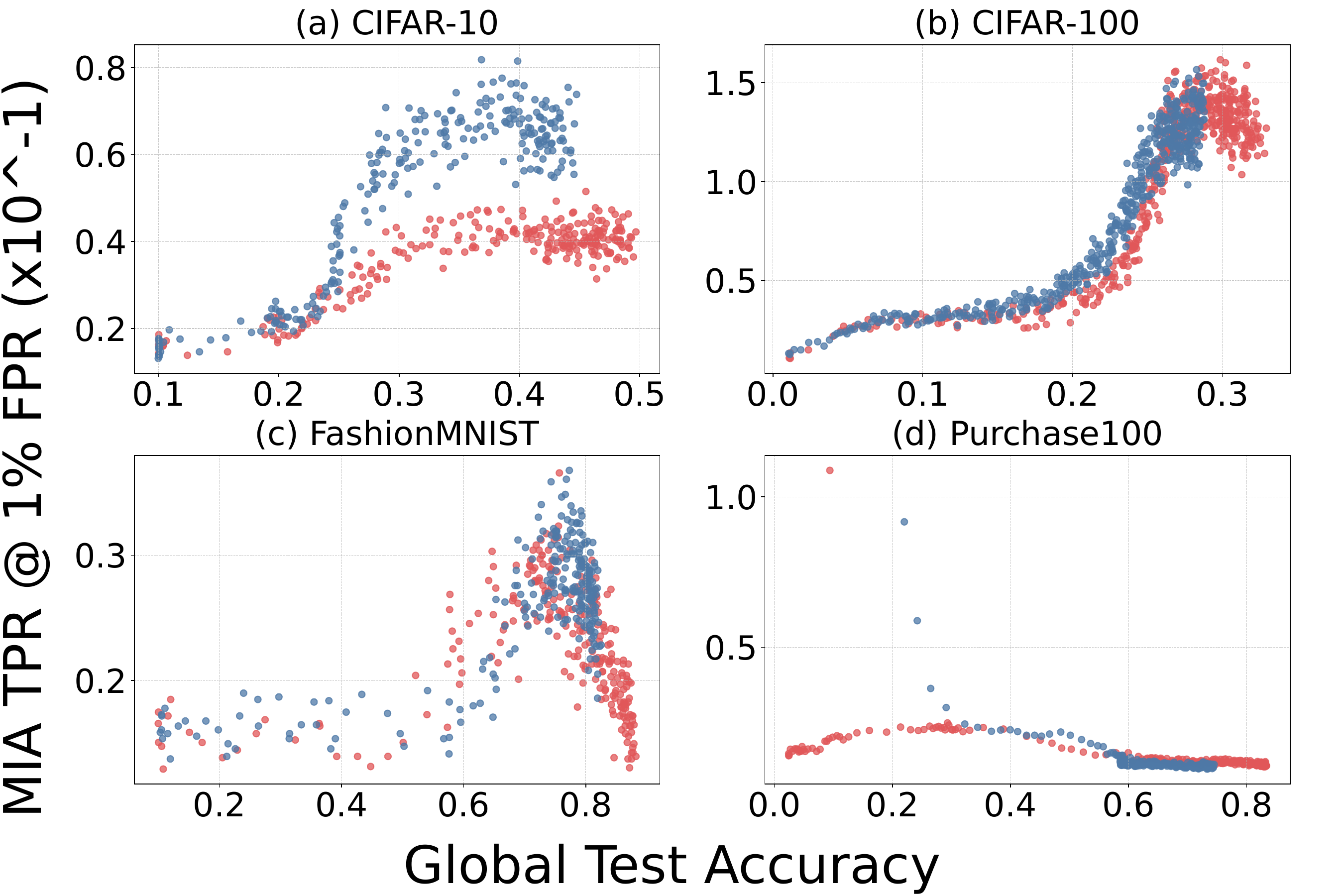}
    \end{minipage}
    \caption{Trade-off between MIA Accuracy and Global Test Accuracy and between MIA \tpratlowfpr{} and Global Test Accuracy across different datasets, comparing Base Gossip and SAMO, on a 5-regular graph with 150 nodes (60 nodes on CIFAR100).}
    \label{fig:tradeoff_iid_comparison_protocol}
\end{figure}

To start our evaluation, we want to emphasize why we propose the protocol SAMO (described in Algorithm~\ref{alg:samo}). 
To do that, we show in Figure~\ref{fig:tradeoff_iid_comparison_protocol} the tradeoff between average MIA vulnerability (accuracy and \tpratlowfpr) and average global test accuracy for models trained on the CIFAR-10, CIFAR-100, Fashion-MNIST, and Purchase100 datasets. 
Each point in the plot represents an evaluation during a given round, with the x-axis showing the average model test accuracy and the y-axis showing the corresponding average MIA vulnerability. 
We set the view size to 5 to ensure a sufficient number of neighbors for SAMO to exhibit distinct behavior from Base Gossip, while still operating within a sparse graph topology.

Our results show that given a target test accuracy the SAMO protocol outperforms Base Gossip in most settings, especially in the regimes where the test accuracy is maximal.
The low accuracy found on datasets such as CIFAR-10 is expected due to the constrained communication topology, observed similarly in other studies \cite{Biswas_2025,devosEpidemicLearningBoosting2023}.
More precisely, with SAMO, the maximum global test accuracy obtained ranges from $35.4\%$ and $88.4\%$ relative to an MIA vulnerability ranging from $62.7\%$ to $92.9\%$ accuracy, and $3.66\%$ to $16.1\%$ for \tpratlowfpr. 
In comparison, in the Base Gossip setting the maximum global test accuracy obtained ranges from $29.9\%$ and $82.6\%$ relative to an MIA vulnerability ranging from $63.5\%$ and $94.1\%$ accuracy, and $3.68\%$ to $15.7\%$ for MIA \tpratlowfpr. 
We explain this advantage by the fact that SAMO favors model mixing, as each user receives models from multiple neighbors and delays the aggregation in order to hide his or her models within more models. 

\begin{takeaway}
    \textbf{Takeaway}: Favoring model mixing by incorporating the SAMO protocol offers a better privacy/utility tradeoff.
\end{takeaway}

Given these results, we favor using the SAMO protocol for the remaining experiments.

\subsection{Comparing Static and Dynamic Settings (RQ2)}
\begin{figure}[ht]
    \begin{minipage}{\columnwidth}
    \centering
    \includegraphics[width=0.5\textwidth]{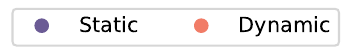}
    \\
    \includegraphics[width=0.90\columnwidth]{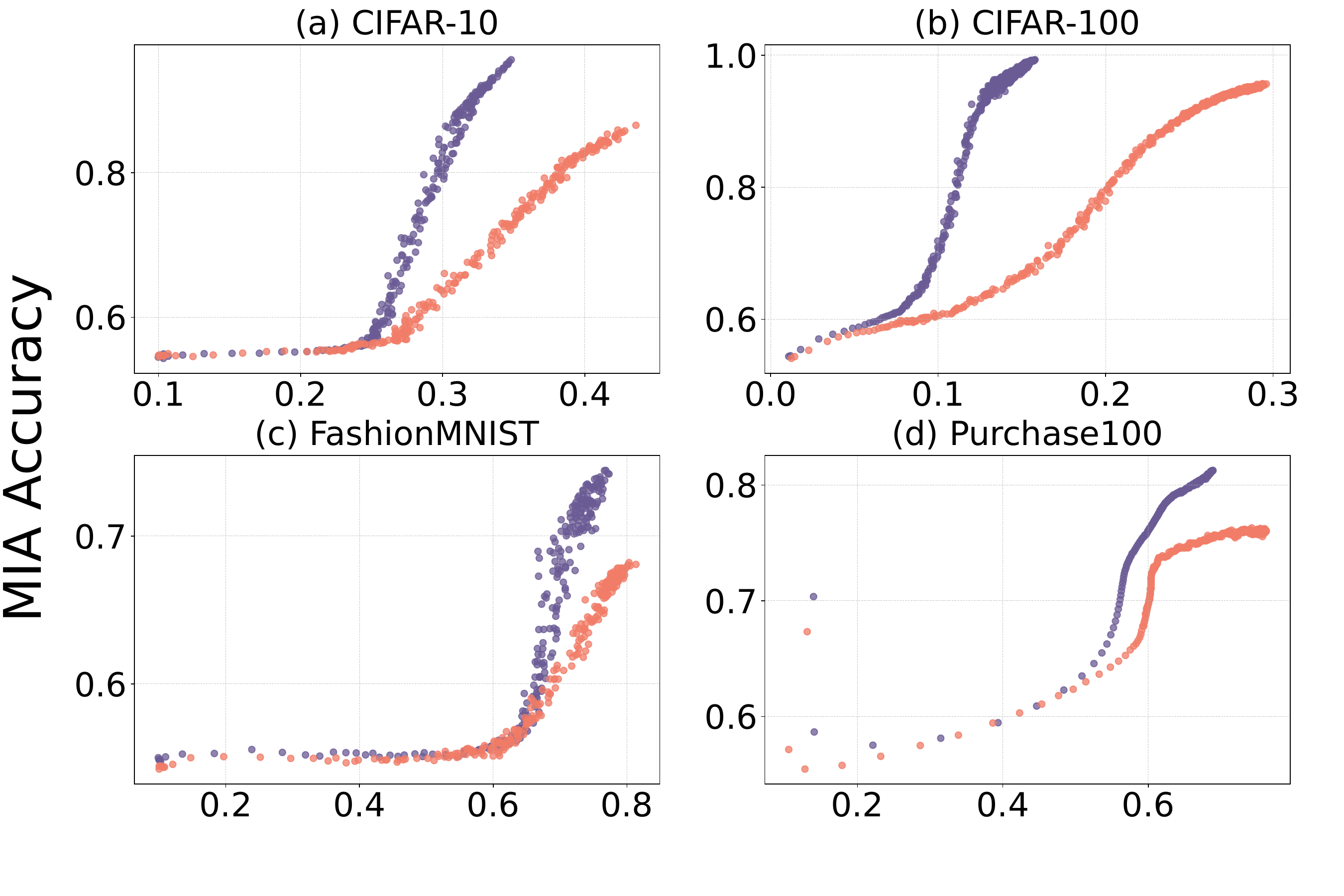}
    \includegraphics[width=0.90\columnwidth]{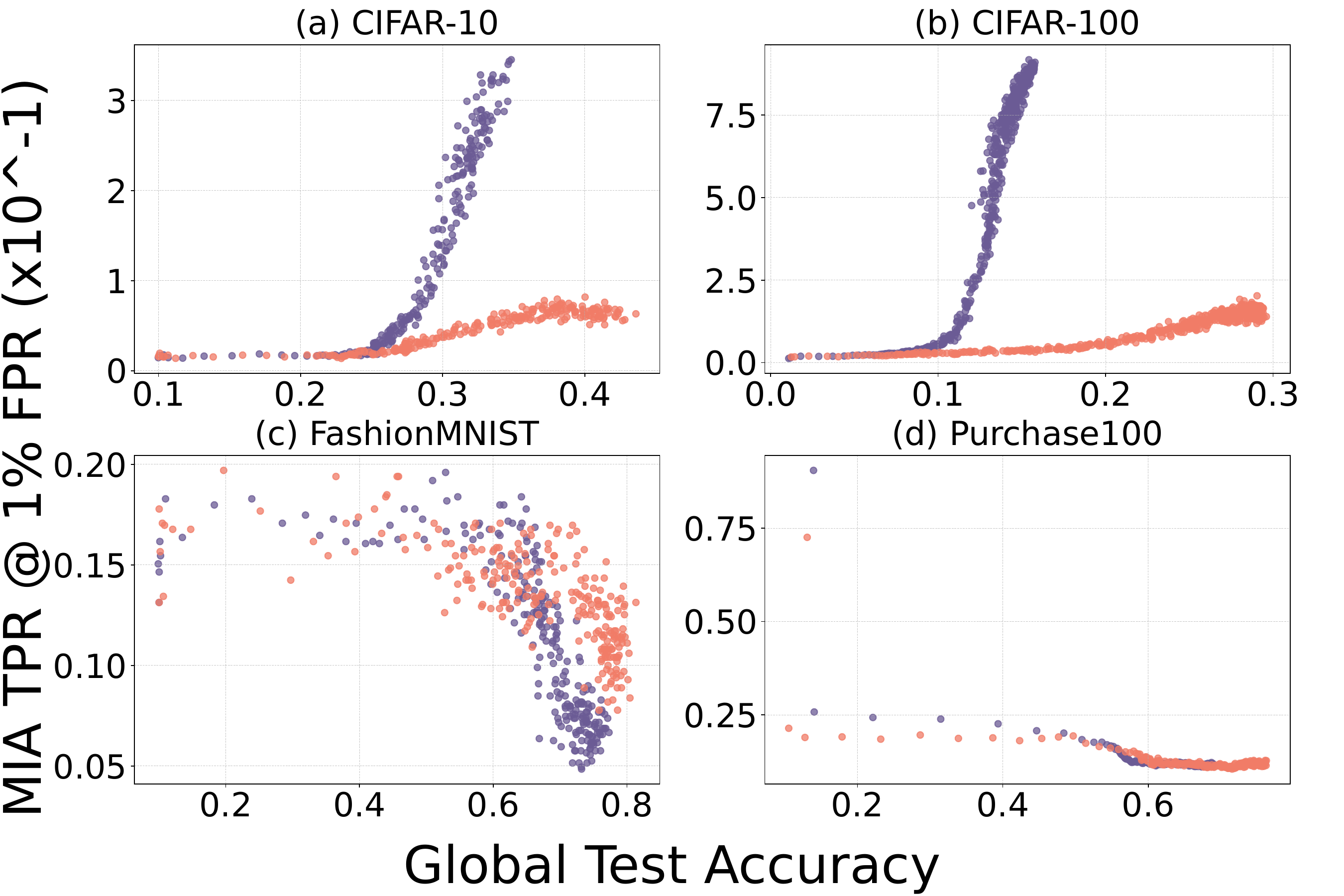}
    \end{minipage}
    \caption{Trade-off between MIA Accuracy and Global Test Accuracy and between  MIA \tpratlowfpr{} and Global Test Accuracy across different datasets, comparing static and dynamic topology setups, on a 2-regular graph with 150 nodes (60 nodes on CIFAR100).}
    \label{fig:tradeoff_iid_comparison_topology}
\end{figure}

Figure~\ref{fig:tradeoff_iid_comparison_topology} shows the trade-off between average MIA vulnerability (accuracy and \tpratlowfpr) and average global test accuracy, comparing static and dynamic topologies. This experiment was conducted on a particularly challenging setting using a sparse graph with a view size of 2.

We observe that in all datasets, settings with dynamic topology outperform the static ones, achieving a better trade-off between MIA vulnerability and global test accuracy. More precisely, in the dynamic setting, the maximum global test accuracy obtained ranges from $30.4\%$ and $80.9\%$ relative to an MIA vulnerability ranging from $67.0\%$ to $94.6\%$ accuracy, and $2.0\%$ to $20.3\%$ for \tpratlowfpr. While, in the static setting, the maximum global test accuracy obtained ranges from $15.8\%$ and $76.4\%$ relative to an MIA vulnerability ranging from $72.0\%$ and $98.8\%$ accuracy, and $2.0\%$ to $91.8\%$ for \tpratlowfpr. 
We observe that models trained with dynamic topologies maintain lower MIA vulnerability at comparable accuracy levels. 

\begin{takeaway}
    \textbf{Takeaway:} Dynamic topologies have a positive impact on both the MIA vulnerability and model utility.
\end{takeaway}

\subsection{On auditing individual privacy with a canary-based approach (RQ3)}
Evaluating individual privacy of worst-case vulnerabilities using standard metrics like TPR at low FPR is often computationally prohibitive, especially in large-scale scenarios involving hundreds of nodes and communication rounds, as in our setup. 
To overcome this, we employ  an efficient alternative called "canaries" as proposed in the work of  ~\citet{aerniTramerMisleading}. 
These crafted data samples simulate worst-case vulnerabilities and are designed to be readily memorized by models lacking privacy safeguards, thus serving as probes for potential leakage. 

Leveraging our network's homogeneity (identical models and parameters per node), we generate canaries by applying a label flipping function to selected training samples and distribute them disjointly and evenly on all the nodes of the communication network. 
We then conduct a targeted, node-specific entropy-based attack on known canary samples, efficiently yielding numerous attack scores for metrics like \tpratlowfpr{}. We used 600 canaries for CIFAR-10/CIFAR-100/FashionMNIST and 1500 for the larger Purchase100 dataset.

\begin{figure}[ht]
    \centering
    \includegraphics[width=0.25\textwidth]{plots/legend_dynamic_static.pdf}
    \begin{minipage}{\columnwidth}
    \centering
    \includegraphics[width=0.90\columnwidth]{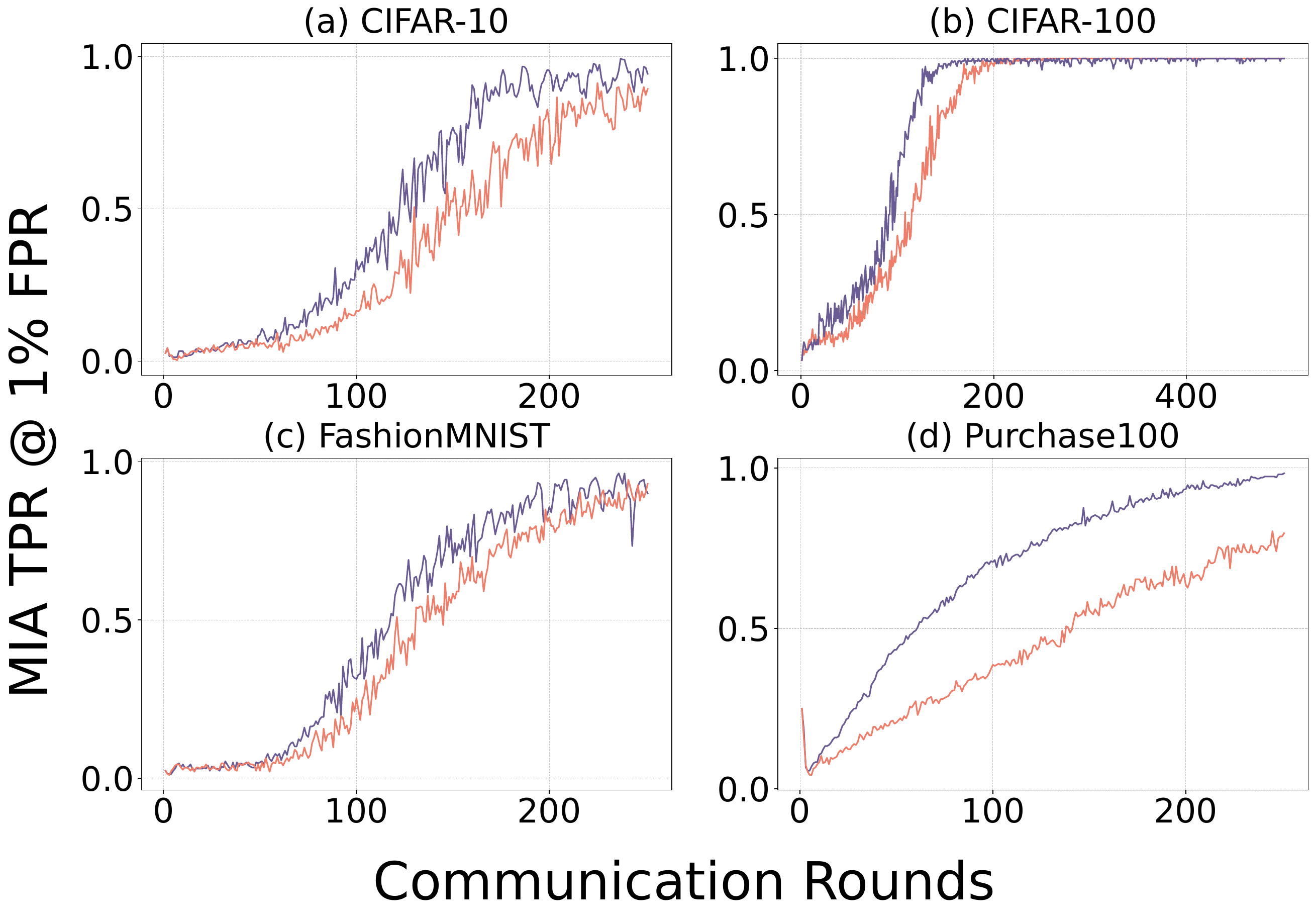}
    \end{minipage}
    \caption{Maximum MIA \tpratlowfpr{} values on the canary set over communication rounds across different datasets, comparing static and dynamic topology setups, on a 2-regular graph with 150 nodes (60 nodes on CIFAR100).}
    \label{fig:mia-vulnerabilities-cifar10-dyn_vs_stat_all_canary}
\end{figure}

Figure~\ref{fig:mia-vulnerabilities-cifar10-dyn_vs_stat_all_canary} presents the evolution of the maximum \tpratlowfpr{} across communication rounds, over 10 runs using distinct canary sets per run.
We can see just how powerful this attack is, displaying results of up to 100\% as these canaries are specially designed to be memorized by node models, making the attacker's task easier.
However, we can see that using a dynamic topology marginally reduces maximum MIA vulnerability in the vast majority of datasets, with a very large reduction in the case of Purchase100.
These experiments reveal that dynamic topologies tend to reduce the efficacy of the attack for outlier cases such as the one simulated here.

\begin{takeaway}
    \textbf{Takeaway:} The positive impact of using a dynamic topology is also observed from the point of view of individual privacy.
\end{takeaway}

\subsection{Impact of the View Size on the Privacy/Utility Tradeoff (RQ4)}

\begin{figure}[ht]   
    \begin{minipage}{\columnwidth}
        \centering
        \includegraphics[width=0.80\columnwidth]{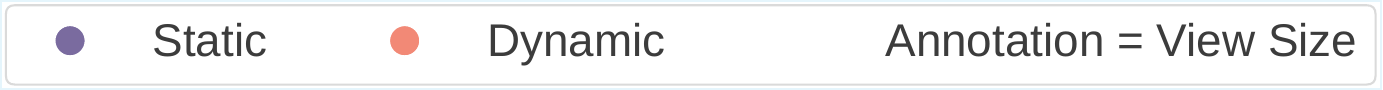}
        \includegraphics[width=\columnwidth]{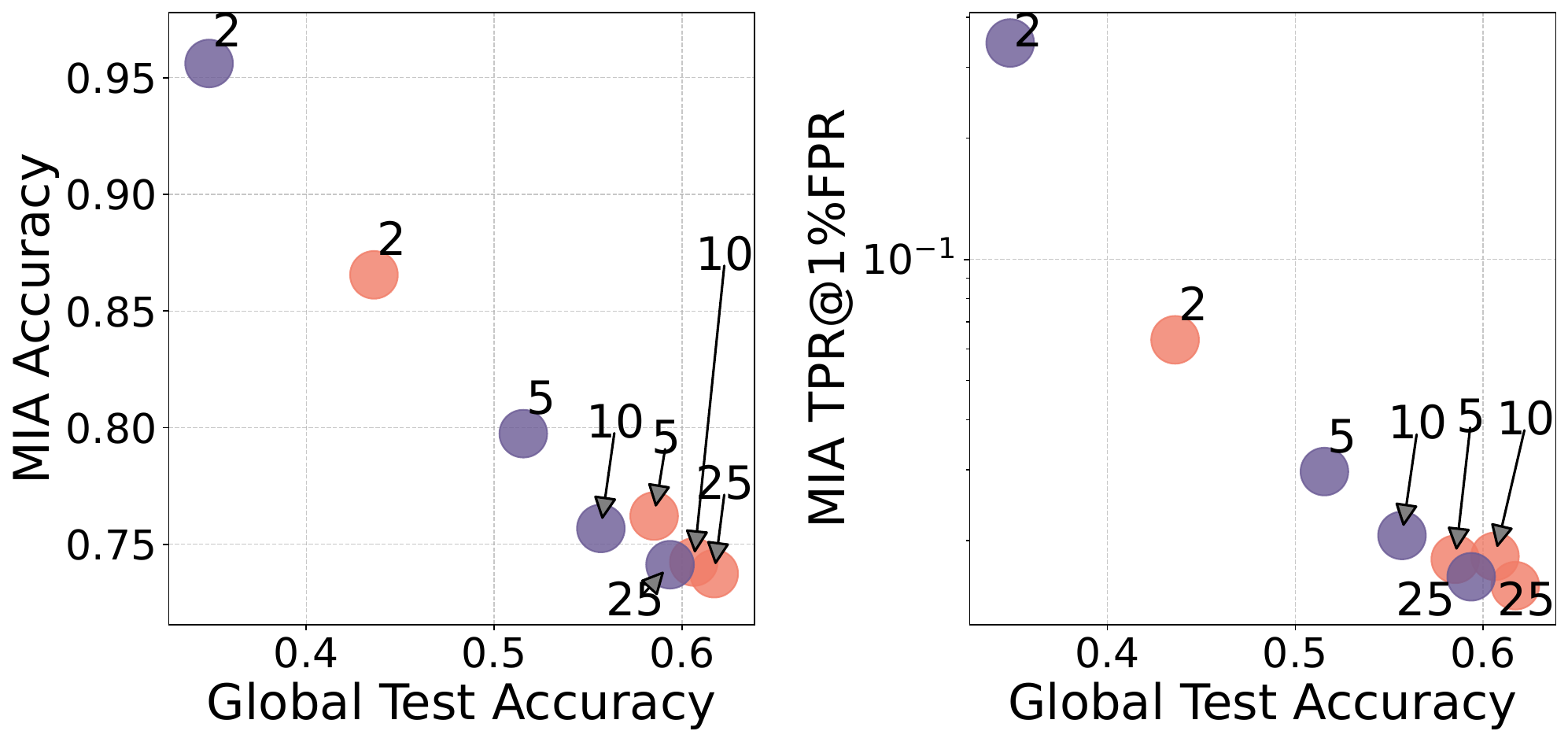}
    \end{minipage}
    
    \caption{Comparison of maximum average MIA accuracy and \tpratlowfpr{} and according global test accuracy across different network view sizes and topology setups on the CIFAR-10 dataset using SAMO, with regular graph with 150 nodes.}
    \label{fig:mia-vulnerabilities-cifar10-view-size}
\end{figure}

In this experiment, we aim to study the impact of the view size on the trade-off between utility and privacy. 
The results are presented in Figure~\ref{fig:mia-vulnerabilities-cifar10-view-size} with the maximum average MIA vulnerabilities (accuracy and \tpratlowfpr) and the maximum average global test accuracy on the CIFAR-10 dataset with various view sizes.
The results show that the dynamic settings outperform the static settings in both utility and privacy. 
Interestingly, upon increasing the view size the differences between dynamic and static gets smaller. This is anticipated as increasing the view size gets the communication graph closer to a complete graph where both settings are equivalent. Also, the increase in view size is overall a strict benefit for the privacy/utility trade-off but this is at the cost of increasing the number of models sent per user (i.e., messages). Overall, profiting from both dynamic topology and increased view size there is a better privacy/utility tradeoff to be found. In particular, using a dynamic setting with a view size of 10 is quasi-similar to a static setting with a view size of 25 (with 2.5 times fewer models sent). 

We conclude that both dynamicity and increased view size have a positive impact on MIA vulnerability as they both favor the model mixing. This intuition is further investigated in Section~\ref{sec:mixing}

\begin{takeaway}
    \textbf{Takeaway:} Increasing the view size closes the gap of dynamic over static but at the cost of more models sent. 
\end{takeaway}

\subsection{Impact of non-i.i.d. Data Distribution  (RQ5)}

In this experiment, our goal is to examine the impact of non-iidness in Gossip Learning. To achieve this, we use the classical method for enforcing a non-i.i.d. setting~\cite{li2022}, sampling from the Dirichlet distribution to apply label imbalance between each node. Specifically, using a Dirichlet distribution with parameter $\beta$, we sample the proportion $p_k \sim Dir_N(\beta)$ of records with label $k$ across the training sets of the $N$ nodes. The parameter $\beta > 0$ here controls the degree of heterogeneity, where a lower value of $\beta$ ($\le 0.1$) yields a higher label imbalance among nodes. In contrast, a higher $\beta$ value yields a distribution closer to the i.i.d. distribution.

\begin{figure}[ht]
    \centering
    \begin{minipage}{\columnwidth}
        \centering
        \begin{subfigure}[b]{0.53\textwidth}
            \centering
           \includegraphics[width=\textwidth]{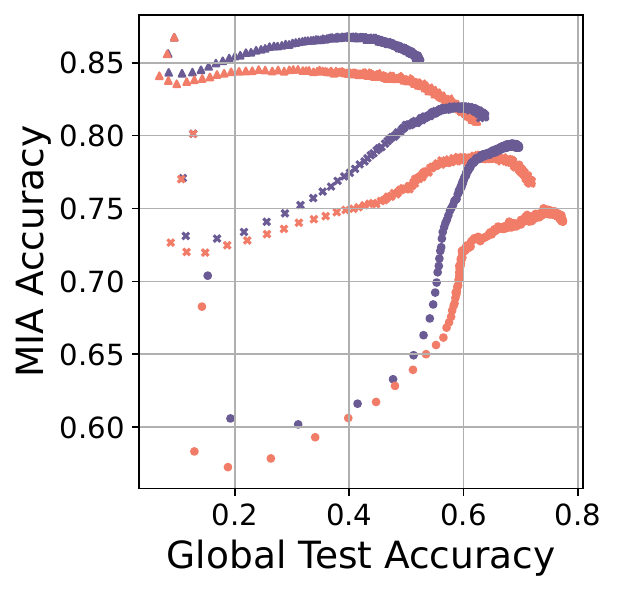}
        \end{subfigure}
        \begin{subfigure}[b]{0.44\textwidth}
            \centering
            \includegraphics[width=\textwidth]{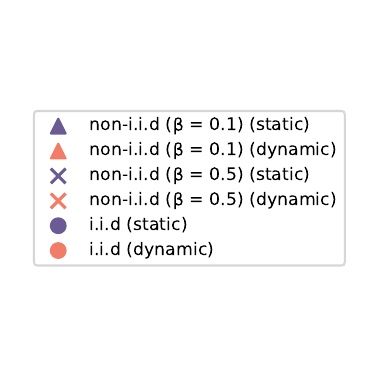}

        \end{subfigure}
    
    \end{minipage}
    \caption{Trade-off between MIA Accuracy and Global Test Accuracy on the Purchase100 with varying degree of heterogeinity (i.i.d, dirichlet $\beta = 0.5$ and $\beta = 0.1$), comparing static and dynamic topology setups, on a 2-regular graph with 150 nodes}
    \label{fig:non-iid}
\end{figure}

The results are presented in Figure~\ref{fig:non-iid}, showing the tradeoff between global test accuracy and MIA vulnerability (accuracy). As expected, the overall global test accuracy decreases with a higher non-i.i.d. setting from a maximum of $77.4\%$ for the i.i.d. setting until a maximum of $52.4\%$ for non-i.i.d. with $\beta=0.1$, as the non-i.i.d. setting is known to make the collaborative learning process more challenging. However, more notably, the MIA vulnerability increases significantly across all training rounds. This is because the learned models tend to align a global model with data distribution close to the overall global distribution, making local updates more prone to leaking membership information.

Regarding the difference between static and dynamic settings, we observe similar conclusions as in the previous experiment, with dynamic settings consistently outperforming static ones. However, the increase in MIA vulnerability due to non-iidness remains prominent regardless of the setting.
Often, the dynamic setting successfully approaches the results of the static setting with lower non-iidness but never completely bridges the gap.  

\begin{takeaway} \textbf{Takeaway:} Non-i.i.d. data distribution among nodes significantly increases MIA vulnerability across all training rounds, and dynamic setting alone is insufficient to completely bridge this gap.
\end{takeaway}

\subsection{Rethinking the Link Between Generalization Error and MIA Vulnerability (RQ6)}
\begin{figure*}[ht]
    \centering
    \includegraphics[width=0.25\textwidth]{plots/legend_dynamic_static.pdf}

    \begin{minipage}{\textwidth}
        \centering
        \begin{subfigure}[b]{0.24\textwidth}
            \centering
           \includegraphics[width=\textwidth]{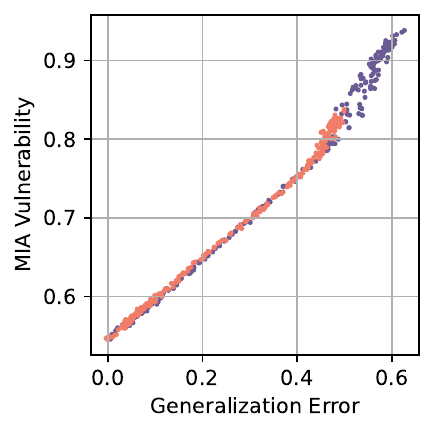}
            \caption{CIFAR-10}
            \label{fig:cifar10_gen_error_tradeoff}
        \end{subfigure}
        \begin{subfigure}[b]{0.24\textwidth}
            \centering
            \includegraphics[width=\textwidth]{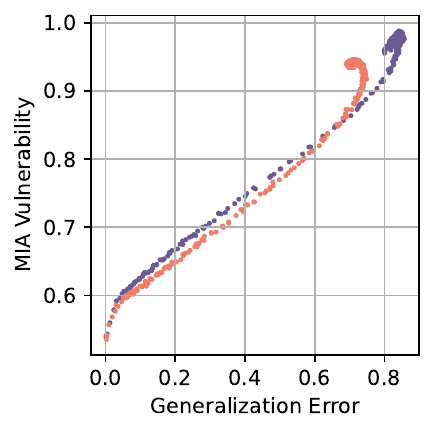}
            \caption{CIFAR-100}
            \label{fig:cifar100_gen_error_tradeoff}
        \end{subfigure}
        \begin{subfigure}[b]{0.24\textwidth}
            \centering
           \includegraphics[width=\textwidth]{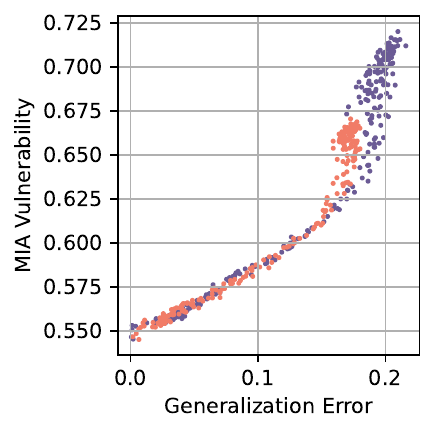}
            \caption{FashionMNIST}
            \label{fig:NICO_gen_error_tradeoff}
        \end{subfigure}
        \begin{subfigure}[b]{0.24\textwidth}
            \centering
           \includegraphics[width=\textwidth]{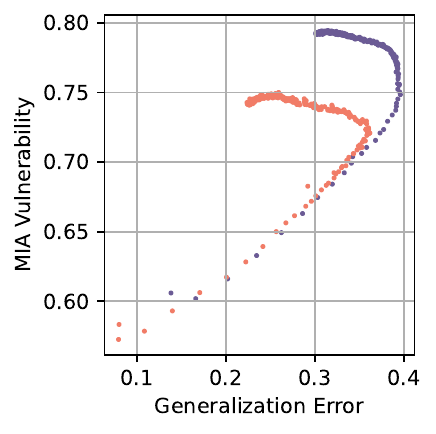}
            \caption{Purchase100}
            \label{fig:Purchase100_gen_error_tradeoff}
        \end{subfigure}
    \end{minipage}

    \caption{Trade-off between MIA vulnerability and Generalization Error across different datasets, comparing Base Gossip and SAMO learning protocols.}
    \label{fig:gen_error}
\end{figure*}

The relationship between generalization error (Equation~\eqref{eq:gen_error}) and MIA vulnerability has been extensively studied in the literature, particularly in the centralized setting~\cite{shokri2017membership, Song2019}. This focus is understandable, as most MIA attacks rely on training overfitting, which is exposed through loss function evaluations on specific examples. The distinction between data points seen during training and those unseen is a crucial factor enabling the separation of members from non-members in MIA attacks. Intuitively, a model that generalizes well should be more resilient to MIA attacks. 
In our setting, however, we are not dealing with a centralized framework involving a single model. Instead, as demonstrated in previous experiments, models undergo various stages of mixing with others trained on different node data. This is why, we investigate the connection between generalization error and MIA vulnerability in our fully decentralized context.
The results, depicted in Figure~\ref{fig:gen_error}, illustrate the tradeoff between average MIA vulnerability (accuracy) and average generalization error. Each point in the plot represents an evaluation during a fixed round, with the x-axis showing the average generalization error and the y-axis showing the corresponding average MIA vulnerability.
The findings reveal dataset-specific behaviors. For example, CIFAR-10 demonstrates an intuitive relationship between generalization error and MIA vulnerability. However, CIFAR-100 and Fashion-MNIST exhibit scenarios where, for a fixed generalization error, MIA vulnerability varies significantly. This phenomenon is even more pronounced for Purchase100, where distinct regimes of MIA vulnerability are observed for the same generalization error.

\begin{figure}[ht]
    \centering
    \includegraphics[width=0.25\textwidth]{plots/legend_dynamic_static.pdf}

    \includegraphics[width=\columnwidth]{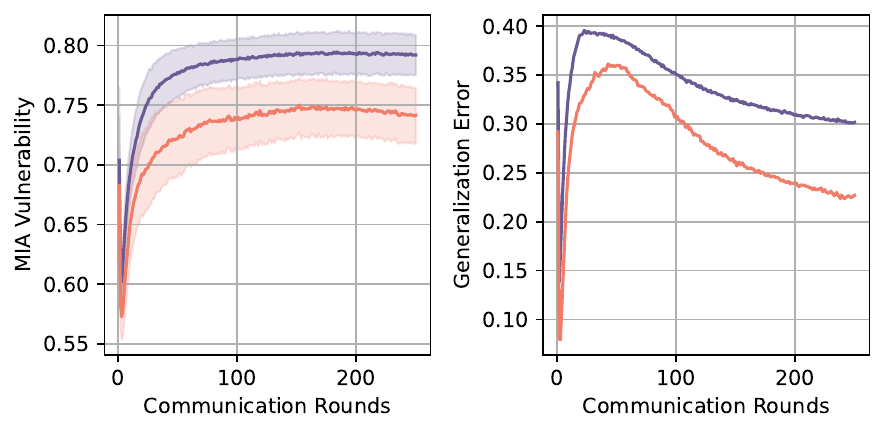}
    \caption{Average MIA Accuracy and Generalization Error over communication rounds on Purchase100 dataset using the Merge Once Protocol, with a 2-regular graph of 150 nodes.}
    \label{fig:gen_error_round}
\end{figure}

To delve deeper, Figure~\ref{fig:gen_error_round} highlights the tradeoff across training rounds. We observe that leakage introduced during an earlier round cannot be mitigated simply by improving the generalization error in subsequent rounds. Once a model overfits, increasing test accuracy and reducing generalization error do not eliminate the existing MIA vulnerability. Our interpretation suggests that the peak generalization error, typically occurring early in the training process, determines the maximum MIA vulnerability observed later.

\begin{takeaway}
    \textbf{Takeaway:} Early overfitting drives persistent MIA vulnerability, which generalization improvements cannot fully mitigate.
\end{takeaway}

This observation aligns with the concept of the Critical Learning Period phenomenon explored in FL \cite{Yan_Wang_Li_2022}. This work emphasizes that changes during the early learning phases significantly impact the best accuracy achievable by the end of training. In our work, we find experimentally a similar pattern, but in terms of MIA vulnerability rather than accuracy.

\subsection{Impact of Dynamicity on ensuring Differential Privacy (RQ7)}

There remains the question of combining these design practices with more direct privacy protection techniques. In this experiment, we study the impact of the dynamicity of the communication topology while Differential Privacy (DP) is ensured. Formally, a stochastic algorithm $\mathcal{M}$ satisfies $(\epsilon, \delta)$-Differential Privacy (smaller parameters correspond to stronger guarantees) if for any two adjacent datasets $D,D'$ and any output $S$, we have $Pr[\mathcal{M}(D) \in S] \le e^{\epsilon}Pr[\mathcal{M}(D') \in S] + \delta$. In our decentralized setting, DP can be enforced on the node-level, by clipping local gradients and adding Gaussian noise with an adequate variance to the clipped gradient at each step, therefore spending a pre-determined privacy budget per step. We use the composition rule of Rényi Differential Privacy (RDP) \cite{renyiDP} to track the overall privacy budget given per-iteration privacy budgets. In practice, our implementation uses Opacus's DP-SGD with RDP accounting. 

\begin{figure*}[ht]
    \includegraphics[width=0.28\textwidth]{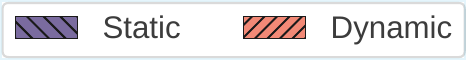}
    \centering
    \begin{minipage}{\textwidth}
    \centering
    \includegraphics[width=\columnwidth]{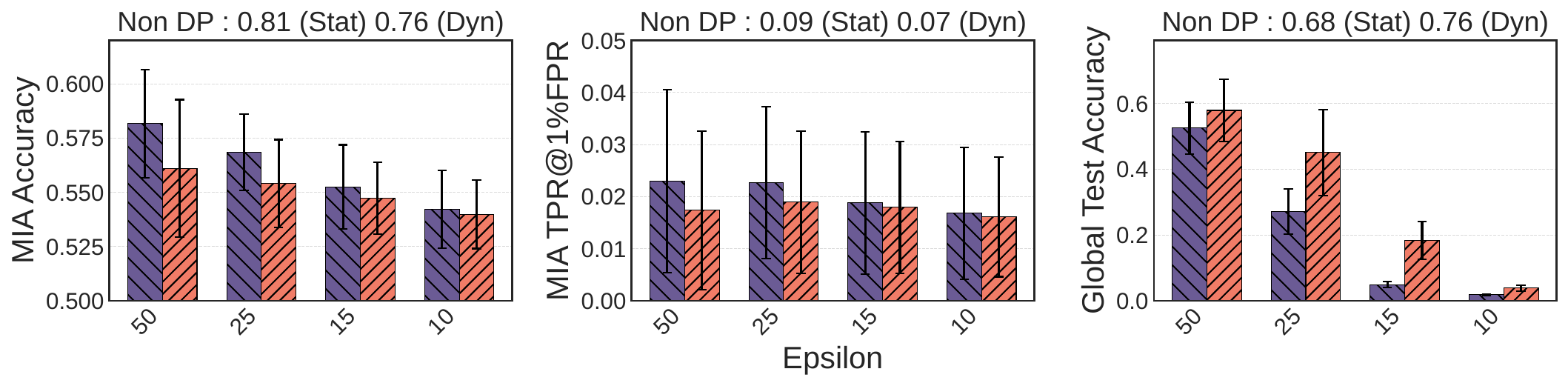}
    \end{minipage}

    \caption{Comparison of maximum average MIA Accuracy and \tpratlowfpr{} with global test accuracy across different DP-SGD privacy budget ($\epsilon$ for $\delta = 10^{-5}$) and topology setups on the Purchase100 dataset using SAMO, with regular graph with 150 nodes.}
    \label{fig:mia-vulnerabilities-purchase100-dp}
\end{figure*}

Figure~\ref{fig:mia-vulnerabilities-purchase100-dp} illustrates the tradeoff between the MIA efficiency and the average utility of the models trained by nodes. The results indicate that applying DP-SGD leads to a decrease in both average model utility and MIA attack efficiency. This effect is particularly pronounced under moderate to stricter DP guarantees (\eg $\epsilon < 25$).

The dynamic setup yields a superior tradeoff between model utility and privacy compared to the static setup, potentially enabling relaxation of the local DP budget. For example, the dynamic setting experiment with $\epsilon=50$ shows much better average model accuracy and lower MIA attack efficiency than the static setting experiment with $\epsilon=25$.

\begin{takeaway}
    \textbf{Takeaway:} While ensuring DP has an impact on model utility, using dynamic topologies yields a better MIA vulnerability/model utility tradeoff.
\end{takeaway}

%% file: content/mixingrate.tex
\section{Mixing with Dynamic Graphs}\label{sec:mixing}

In Section~\ref{sec:exp}, we demonstrated the benefits of dynamic settings and increased view sizes in improving both privacy and utility. Our intuition is that the mixing properties of the learning protocol play a crucial role in reducing MIA vulnerability. Therefore, in this section, we aim to provide a deeper understanding of how dynamic settings and larger view sizes enhance the mixing rate of models.
To investigate the mixing properties, we leverage existing findings on the influence of the spectral properties of the weighted adjacency matrix associated with gossip exchanges \cite{boyd2006randomized}. To conduct the analysis, we assume an idealized scenario with synchronous iterations where all nodes simultaneously wake up to exchange their models and no local updates are performed.

The simplified analysis considers essentially a consensus task. Let  $\iterCnt \in \mathbb{N}$ be a fixed number of iterations. At each (synchronous) iteration $\iter \in  \{1,\dots, \iterCnt\}$,  each node $i \in \partySet$ performs the following mixing step:
\begin{equation} 
\model{\iter+1}{i} \gets \frac{1}{\neighSize+1}\sum_{j \in \neighbors{\iter}{i} \cup \{i\}} \model{\iter}{i}
\label{eq:mixingstep} 
\end{equation}
Where $\model{\iter}{i} \in \modelDom$ is the model of node $i$ at iteration $\iter$ and $\model{0}{i}$ is equal to the initial model  $\modelASynch{0} \in \modelDom$. 
While it is already evident that graphs $\graph{1}, \dots \graph{\iterCnt}$ defined by neighborhoods $\neighbors{\iter}{i}$ for all $i \in \partySet$ and $\iter \in \{1,\dots, \iterCnt\}$ impact convergence of consensus, this is also holds for decentralized learning \cite{koloskova2020unified,even2024asynchronous}. 

We now analyze mixing properties. 
For any iteration $\iter \in \{1,\dots,\iterCnt \}$, let  $\modelVectorIt{\iter} \in \modelDom^n$ be the vector whose elements are the models of each node, and $\mixMatrixIt{\iter} \in \mathbb{R}^{n\times n}$ be the weighted adjacency matrix  (or mixing matrix) of the graph used at iteration $\iter$. Specifically, for our $k$-regular graphs $\mixMatrixItEl{\iter}{i}{j} = 1/(\neighSize+1)$ if and only if $i$ and $j$ are neighbors or $j = i$, and $\mixMatrixItEl{\iter}{i}{j}=0$ else. 

Exchanges at Equation \eqref{eq:mixingstep} can be summarized by 
\begin{equation}
\modelVectorIt{\iter+1} \gets \mixMatrixIt{\iter} \modelVectorIt{\iter}. 
\end{equation} 
It has been proven in the work of \citet{boyd2006randomized} that if a mixing matrix $\mixMatrix \in \mathbb{R}^{n\times n}$ is symmetric and doubly stochastic (that is, the sum of the elements of any column or row are equal to $1$), then for any vector $\vect \in \mathbb{R}^n$  we have that
\begin{equation}
	\lTwoNorm{  \mixMatrix \vect - \oneVec \avgVec  } \le \secondEig{\mixMatrix} \lTwoNorm{ \vect - \oneVec \avgVec }, 
	\label{eq:mixingrate}
\end{equation} 
where $\avgVec = \frac{1}{n} \sum_{i \in \partySet} \vect_i$ is the average of the elements of the vector,  $\secondEig{\mixMatrix}$ is the second largest eigenvalue of $\mixMatrix$ and $\oneVec$ is the vector of all ones. Essentially,  \cite{boyd2006randomized} shows that $\secondEig{\mixMatrix}$ quantifies the reduction of the distance between values (in our case, models \footnote{For simplicity, we perform our analysis treating models as scalars. However, if we consider models as vectors of scalars (i.e., $\modelDom = \mathbb{R}^d$ where $d$ is the number of parameters of the model), the current analysis holds for each model parameter.}) of each node and the average of all values. The smaller $\secondEig{\mixMatrix}$, the better the mixing. Given that we use undirected $k$-regular graphs and the weights of the exchanges is equal for all edges, we have that for all $\iter \in \{1, \dots, \iterCnt\}$, $\mixMatrixIt{\iter}$ is doubly stochastic and symmetric and hence we can use Equation \eqref{eq:mixingrate} in our simplified consensus protocol.

We focus on the mixing properties of the matrix $\mixSequence$ obtained by performing the matrix product 
\(
\mixMatrixIt{\iterCnt} \dots \mixMatrixIt{1}.
\)
For any vector $\vect \in \mathbb{R}^n$, 
$\mixSequence \vect$ is the result of performing the sequence of mixing exchanges determined by $\mixMatrixIt{1}$,$\dots$, $\mixMatrixIt{\iterCnt}$ over $\vect$. 

An immediate bound given by 
\begin{equation}
	\lTwoNorm{  \mixSequence \vect - \oneVec \avgVec  } \le \prod_{\iter=1}^\iterCnt \secondEig{\mixMatrixIt{\iter}} \lTwoNorm{\vect - \oneVec \avgVec }.
	\label{eq:mixingSec} 
\end{equation}
can be obtained from sequentially applying Equation \eqref{eq:mixingrate} to each factor $\mixMatrixIt{\iter}$ of $\mixSequence$. However, Equation \eqref{eq:mixingSec} only takes into account individual mixing properties of each $\mixMatrixIt{\iter}$ in isolation,  ignoring the effects of varying the communication graph. 

Since  \(\mixSequence\) is also doubly-stochastic when   $\mixMatrixIt{1}$,$\dots$, $\mixMatrixIt{\iterCnt}$ are doubly-stochastic, a more accurate bound can be obtained by directly applying that Equation \eqref{eq:mixingrate} using $\secondEig{\mixSequence}$. This captures the overall mixing properties of the graph sequence. In the \emph{static setting},  $\mixMatrixIt{\iter} = \mixMatrix$  for all $\iter \in \{1,\dots, \iterCnt \}$ and some $\mixMatrix$ initially chosen. 
In the \emph{dynamic setting} all nodes are randomly permuted at each iteration. It is easy to see that for the static case $\secondEig{\mixSequence} = \secondEig{\mixMatrix}^\iterCnt$, while in the dynamic case the closed form of $\secondEig{\mixSequence}$ is not evident.

\begin{figure}
	\includegraphics[scale=0.4]{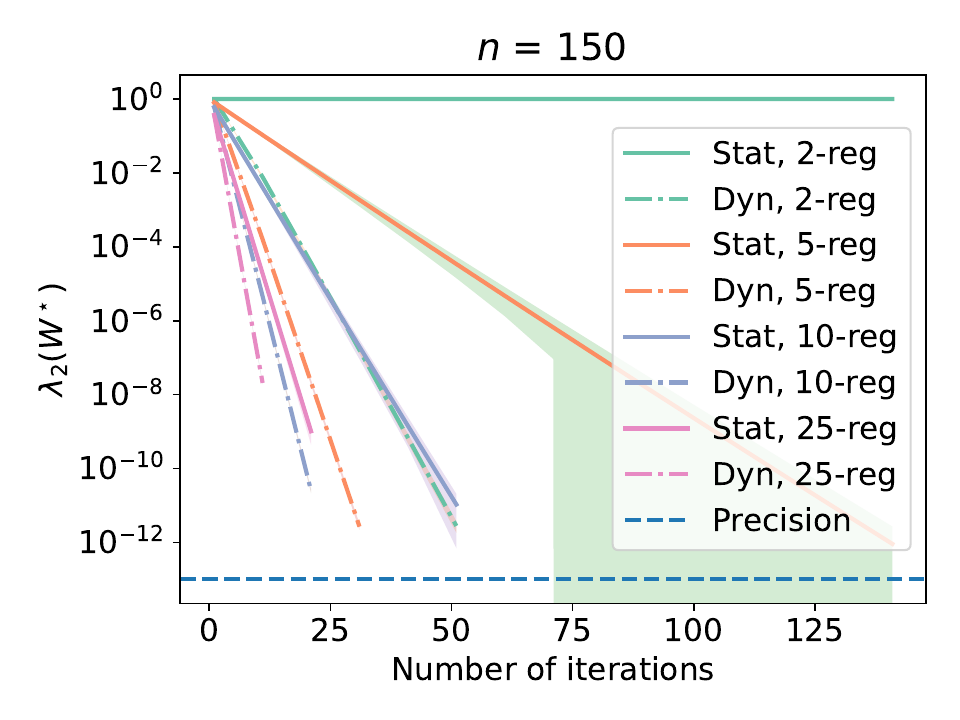}
	\caption{Values of  $\secondEig{\mixSequence}$ depending on the number of iterations for $k$-regular graphs ($k$-reg) of different degrees in the dynamic (Dyn) and static (Stat) settings.}
	\label{fig:mixing}
\end{figure}

In Figure \ref{fig:mixing}, we empirically show how $\secondEig{\mixSequence}$ decreases as iterations increase. We show it for the same topologies used in Section \ref{sec:exp}. 
All points in the curve are reported over an average of 50 runs of gossip simulation together with their standard deviation. We can see that for the same degree $\neighSize$, $\secondEig{\mixSequence}$ decreases much faster for dynamic graphs than for static graphs. Moreover, the standard deviation is negligible in the dynamic case, indicating that bad mixing scenarios occur with negligible probability. 

From this analysis, we can better understand why the dynamic setting is favorable for reducing MIA vulnerability. Models in the dynamic setting tend to align more closely with consensual models and exhibit less bias toward the peculiar model of an individual client, which would otherwise increase MIA vulnerability.

%% file: content/discussion-summary.tex
\section{Summary of Takeaways and Actionable Recommendations\label{sec:summary}}

Our experimental analysis provides several important insights into the factors influencing MIA vulnerability and utility in Gossip learning. Below, by summarizing the above takeaways, we offer actionable recommendations for designing privacy-aware decentralized learning architectures:

\begin{itemize}
\item \textbf{Static vs Dynamic Graph Topologies:}
From our study, we recommend using dynamic graph structures to improve model mixing and mitigate privacy risks, especially as the communication network becomes large and sparse. These topologies should be paired with robust peer-sampling protocols.
\revision{Furthermore, they complement privacy preserving techniques such as DP-SGD since dynamic graph topologies reduce the negative impact on model utility and decrease the vulnerability to MIA attacks.}

\item \textbf{Model Mixing:}
Our study shows that incorporating advanced mixing strategies like SAMO reduces individual local biases in models and achieves better privacy-utility tradeoffs.

\item \textbf{View Size:}
While increasing view size is generally beneficial, it should be balanced against the increased communication cost. For practical implementations, we recommend considering dynamic settings with moderate view sizes to maximize efficiency.

\item \textbf{Addressing Non-i.i.d. Data:}
Our study shows the high impact of non-i.i.d. data distribution which asks for stronger protection mechanisms. We also recommend exploring adaptive learning protocols that account for data heterogeneity, such as adapting the ones introduced in FL~\cite{ye2023heterogeneous}. For instance, using nodes pre-clustering~\cite{briggs2020federated} or knowledge distillation using public data~\cite{fang2022robust}. 
\item{\textbf{Early Overfitting:}}
From our study, we recommend focusing on strategies to prevent early overfitting, such as regularization, dynamic learning rates, or delayed aggregation, to limit the persistent impact of initial vulnerabilities.
\end{itemize}

By implementing these recommendations in completion to privacy-preserving mechanisms (see Section~\ref{sec:related:defense}), decentralized learning systems can achieve a more favorable balance between utility and privacy, paving the way for safer and more effective collaborative learning environments.

%% file: content/related_works.tex
\section{Related Works}\label{sec:related}
\subsection{Decentralized Learning}
\revision{
While the literature on decentralized learning is significantly exhaustive, it shares one common objective: tackling the drawbacks of Federated Learning that stem from its centralized nature (\eg, single point of failure, governance, scalability \ldots). The first such attempts took the form of semi-decentralized FL. For instance,~\cite{guo2021prefer} assumes the presence of several servers while~\cite{lin2021semi} envisions a hierarchical workflow consisting of several clusters of clients. While these approaches mitigate some of the previously mentioned drawbacks, they do not fully address issues related to governance and trust. To tackle these concerns, subsequent works have turned towards fully decentralized learning. Recent surveys~\cite{yuan2024decentralized, beltran2023decentralized} distinguish these approaches based on (1) the roles played by nodes in the protocol and (2) the communication scheme, both in terms of topology and synchronicity. Regarding the first point, most of the literature centers around the classical trainer and aggregator roles. In fully decentralized settings such as ours, clients are traditionally assumed to play both roles. This setup is often considered appealing as it enables nodes to retain full control over the training and aggregation processes, including their frequency, allowing them to adapt to their local resources~\cite{zehtabi2022decentralized}. However, some lines of work also consider the proxy role~\cite{kalra2023decentralized}, while works leveraging blockchain systematically introduce a verifier role~\cite{qu2022blockchain} that is relevant for resilience purposes.}

\revision{
While such additional layers are undoubtedly relevant for bridging the gap between theory and practical implementations, they do not significantly impact the privacy of decentralized learning and can always be integrated on top of protocols like ours. Secondly, the communication scheme varies considerably across frameworks. In terms of topology, most of the literature has focused on fully connected graphs due to their convergence properties, which are equivalent to FL~\cite{georgatos2022efficient,xiao2021fully}, and their resilience to network dynamics (\eg, churn and failures). However, such graphs incur quadratic communication costs~\cite{bellet2022d}, making them impractical for large-scale learning and, to some extent, defeating the purpose of decentralization. As a result, there have been increasing efforts toward learning over sparser graphs~\cite{hua2022efficient,guo2025harmonizing}. In this context, Wu et al.~\cite{wu2024topology} recently demonstrated that d-regular expander graphs are near-optimal both in terms of convergence and generalization. Since building d-regular expanders without global knowledge is non-trivial, dynamic regular graphs are often used instead, an approach we adopt in this work. In fact, such an approach has been shown to achieve graph-size-independent consensus rates~\cite{songCommunicationEfficientTopologiesDecentralized2022} (See Section~\ref{sec:related:dynamic} for details). Finally, another critical aspect of the communication scheme relates to synchronicity. Much of the literature assumes settings that enforce synchronization points. While this is easier to model and thus attractive from a theoretical standpoint~\cite{koloskova2020unified}, it is suboptimal in real-world, resource-heterogeneous settings, as it introduces unnecessary idle time and slows convergence~\cite{zehtabi2022decentralized,cao2021hadfl}. In contrast, asynchronism; a core principle of what is commonly termed gossip learning, aligns well with real-world constraints, prompting recent efforts to establish its theoretical foundations~\cite{even2024asynchronous}. It is from this principle that, in this work, we consider an asynchronous, fully decentralized learning setting, which accounts for practical requirements as well as the emerging necessary theoretical grounds.
}
\subsection{Privacy in Decentralized and Gossip Learning.} 
From location-hiding~\cite{gotfryd2017location} to source anonymity~\cite{bellet2020started,jin2023privacy}, The inherent privacy properties of gossip protocols have captivated the research community over the past decade. However, studying how these properties translate to a learning protocol remains relatively under-explored. To the best of our knowledge, ~\citep{pasquini_security_2023} is the first endeavor in this direction. In this work, the authors demonstrated that decentralized learning is at least as vulnerable as its federated counterpart to reconstruction and membership inference attacks. However, their approach relies on partial model aggregation, which we show leads to worse model mixing and, consequently, to more vulnerable models. Moreover, their results rely on assumptions such as adversaries possessing power equivalent to an FL parameter server in some contexts, or the notion that honest nodes have no honest neighbors within a two-hop radius (~\citep{pasquini_security_2023}, Section 4). 

These assumptions are particularly unrealistic for dynamic networks such as ours. Additionally, important characteristics of gossip learning, such as asynchronous communications and the presence of a random peer sampling protocol, are not considered. The threat model of~\citet{pasquini_security_2023} was recently extended in~\citep{mrini2024privacy}, where the authors consider adversaries operating beyond their immediate neighborhoods. While their findings are compelling, they remain limited to static and synchronous settings and further assume i.i.d. data distributions. Concurrently, a recent study~\cite{ji2024re} examines the vulnerability of decentralized learning through a mutual information framework. While their theoretical results regarding the resilience of decentralized learning align with our findings, their analysis is constrained. Specifically, they exclusively consider an i.i.d. setting, with static topologies and synchronous communications. Our work alleviates the assumptions made in these works.

\subsection{Defenses and Protection Mechanisms\label{sec:related:defense}} 
Numerous attempts have been made to enhance privacy in decentralized/gossip learning using various mechanisms. For instance, \citet{cyffers2022privacy} proposes a novel formulation of differential privacy~(DP) that accounts for the position of nodes within the network.  In such an approach, nodes send noisy model updates with formal privacy guarantees. Then \citet{sabater2022accurate} considers a stronger model where adversarial parties can collude and share their view. Moreover, it substantially reduces the perturbations required to achieve privacy using a correlated noise approach. While DP-based solutions are holistic from several perspectives (\eg, adversary knowledge, types of attacks), they often result in significant utility loss~\cite{carey2024evaluating, kim2021federated} and can impede other robustness objectives~\cite{Guerraoui2021DifferentialPA}.

Other works explore secure multi-party computation primitives. For example, \citet{dekker2023topology} proposes a framework where each node performs a sequence of secure aggregations with its neighbors. Although effective, these solutions are often expensive to adapt to more general settings (\eg, Byzantine settings~\cite{so2020byzantine}). Finally, another line of research investigates hardware-based approaches, namely, Trusted Execution Environments~(TEEs). For instance, GradSec~\cite{aitmessaoud2021} prevents privacy attacks in FL by incorporating sensitive model layers within secure compartments. However, this approach has demanding hardware requirements and has been shown to be vulnerable to side-channel attacks~\cite{jiang2022challenges}. Overall, while we consider each of these research directions promising, they are orthogonal to our work, which focuses specifically on understanding the vulnerabilities of models in gossip learning, particularly in relation to dynamicity and mixing rate.

\subsection{Studying Dynamic Communication Topologies}~\label{sec:related:dynamic}
There are mainly two lines of work that investigate dynamic topologies in gossip learning: (i) works that attempt to quantify the impact of dynamicity on the mixing rate and convergence properties, and (ii) works that propose practical protocols leveraging peer sampling algorithms~\cite{belalPEPPEREmpoweringUserCentric2022, hegedusDecentralizedLearningWorks2021, hegedHus2019gossip}. In the mixing rate and convergence line of work, several studies have examined the use of dynamic topologies in decentralized learning, with the majority focusing on the synchronous case~\cite{nedicDistributedOptimizationTimevarying2014, songCommunicationEfficientTopologiesDecentralized2022}. A noteworthy example is Epidemic Learning~\cite{devosEpidemicLearningBoosting2023}, which employs a randomly formed communication topology where, in each round, a node transmits its model to a randomly selected set of peers without coordination between them. The study showed that randomness in peer sampling positively impacts convergence speed, aligning with previous results~\cite{songCommunicationEfficientTopologiesDecentralized2022}. 

Corroborating these findings, many practical peer-sampling protocols have been proposed. For instance, \citet{hegedusDecentralizedLearningWorks2021} demonstrate empirically that gossip learning, enhanced with a peer sampling protocol, can systematically compete with federated settings across various configurations and hyperparameters (e.g., message loss, data distributions). Similarly, \citet{belalPEPPEREmpoweringUserCentric2022} observe that a carefully designed peer sampling protocol can improve model personalization. 
Overall, while our work aligns to some extent with both research directions, it introduces a novel perspective by establishing a crucial connection between the mixing properties of a practical dynamic protocol and its vulnerability to membership inference attacks. To the best of our knowledge, such a connection has not been explored in prior research.

%% file: content/conclusion.tex
\section{Conclusion}\label{sec:conclusion}

In this work, we presented a comprehensive study of MIA vulnerabilities in decentralized learning systems. By analyzing the interplay between model mixing properties, graph dynamics, and data distributions, we provided new insights into the privacy risks inherent in these systems.

Our findings revealed that mixing properties are critical in shaping MIA vulnerabilities. Dynamic graph topologies, coupled with increased view sizes, enhance mixing rates, leading to better privacy-utility tradeoffs. However, the effectiveness of these approaches is constrained when faced with non-i.i.d. data distributions, which amplify MIA risks across all training rounds. Importantly, we demonstrated that early overfitting creates persistent vulnerabilities that cannot be fully mitigated by improving generalization later in the training process.

These insights underline the need for careful design in decentralized learning protocols. By favoring dynamicity and improving model mixing, systems can achieve a better balance between utility and privacy. Future work should focus on developing adaptive mechanisms to address the challenges posed by heterogeneous data distributions and early overfitting, as well as exploring complementary defense techniques, such as differential privacy and secure multi-party computation.

Through our analysis and experiments, we provide actionable recommendations for building more privacy-resilient decentralized learning systems, paving the way for safer collaborative learning environments.